\documentclass[letterpaper,journal]{IEEEtran}

\usepackage{amsmath,amsfonts,amssymb,bm}
\usepackage{array}
\usepackage{booktabs}
\usepackage{cite}
\providecommand{\citep}[1]{\cite{#1}}
\usepackage{graphicx}
\usepackage{multicol}
\usepackage{placeins}
\usepackage[caption=false,font=normalsize,labelfont=sf,textfont=sf]{subfig}
\usepackage{textcomp}
\usepackage{xcolor}
\usepackage{url}
\usepackage[hidelinks,hypertexnames=false]{hyperref}
\renewcommand{\topfraction}{0.85}
\renewcommand{\dbltopfraction}{0.80}
\renewcommand{\textfraction}{0.15}
\renewcommand{\floatpagefraction}{0.85}
\renewcommand{\dblfloatpagefraction}{0.85}
\makeatletter
\newcommand{\wfinlinefloats}{\renewenvironment{figure}[1][]{\par\begingroup\def\@captype{figure}}{\par\endgroup}\renewenvironment{figure*}[1][]{\par\begingroup\def\@captype{figure}}{\par\endgroup}\renewenvironment{table}[1][]{\par\begingroup\def\@captype{table}}{\par\endgroup}\renewenvironment{table*}[1][]{\par\begingroup\def\@captype{table}}{\par\endgroup}}

\makeatother

\newcommand{\vcache}{v-cache}

\def\BibTeX{{\rm B\kern-.05em{\sc i\kern-.025em b}\kern-.08em
    T\kern-.1667em\lower.7ex\hbox{E}\kern-.125emX}}

\begin{document}

\title{WorldForge V2: Eliciting Camera-Trajectory Alignment in Video Diffusion Models}

\author{Chenxi~Song, Yanming~Yang, Tong~Zhao, Ruibo~Li,\\
Joey~Tianyi~Zhou,~\IEEEmembership{Senior Member,~IEEE,} and Chi~Zhang,~\IEEEmembership{Member,~IEEE}\thanks{C. Song, Y. Yang, T. Zhao, and C. Zhang are with Westlake University, Hangzhou, China. R. Li is with Beihang University, Beijing, China. J. T. Zhou is with Singapore Management University, Singapore. Corresponding author: Chi Zhang (e-mail: \href{mailto:chizhang@westlake.edu.cn}{chizhang@westlake.edu.cn}).}}

\markboth{WorldForge V2}{Song \MakeLowercase{\textit{et al.}}: WorldForge V2}

\maketitle

\begin{abstract}
    Video diffusion models have learned rich 3D/4D priors, making them promising for scene generation and video re-rendering. However, their use in these spatial tasks is limited by imprecise camera control, spatial and temporal inconsistencies, and the entanglement of camera motion with scene dynamics. Existing approaches, such as task-specific fine-tuning and post-process warping, often introduce visual artifacts, generalize poorly, or incur high computational costs. We propose WorldForge V2, which aims to achieve accurate camera control while preserving the generative capabilities of pretrained video models. We first develop a training-free guidance framework using warped references to iteratively refine generation along a target camera path. Optical-flow analysis identifies motion-related latent channels for selective guidance, helping separate motion control from appearance. A dual-path strategy further compares guided and unguided predictions to correct artifacts from imperfect references. Together, these components guide viewpoint changes while preserving scene appearance and object motion. To reduce refinement costs, \vcache{} approximates the correction using the preceding denoising prediction, avoiding an additional model evaluation. We further extend motion localization from latent channels to network blocks and introduce Trajectory LoRA, which concentrates adaptation on blocks most responsive to trajectory guidance. A two-stage training strategy teaches these adapters to denoise the states encountered during guided generation, using limited supervision effectively. This enables reliable camera control with very little training data, with a single training scene already supporting generalization to unseen scenes. Quantitative and qualitative evaluations across multiple datasets show that our method achieves state-of-the-art generation quality and trajectory accuracy in camera-controlled video generation.

    \end{abstract}

\begin{IEEEkeywords}
Camera control, video diffusion models, 3D/4D generation.
\end{IEEEkeywords}

\section{Introduction}
\label{sec:introduction}

\begin{figure}[!t]
    \centering
    \includegraphics[width=0.90\columnwidth]{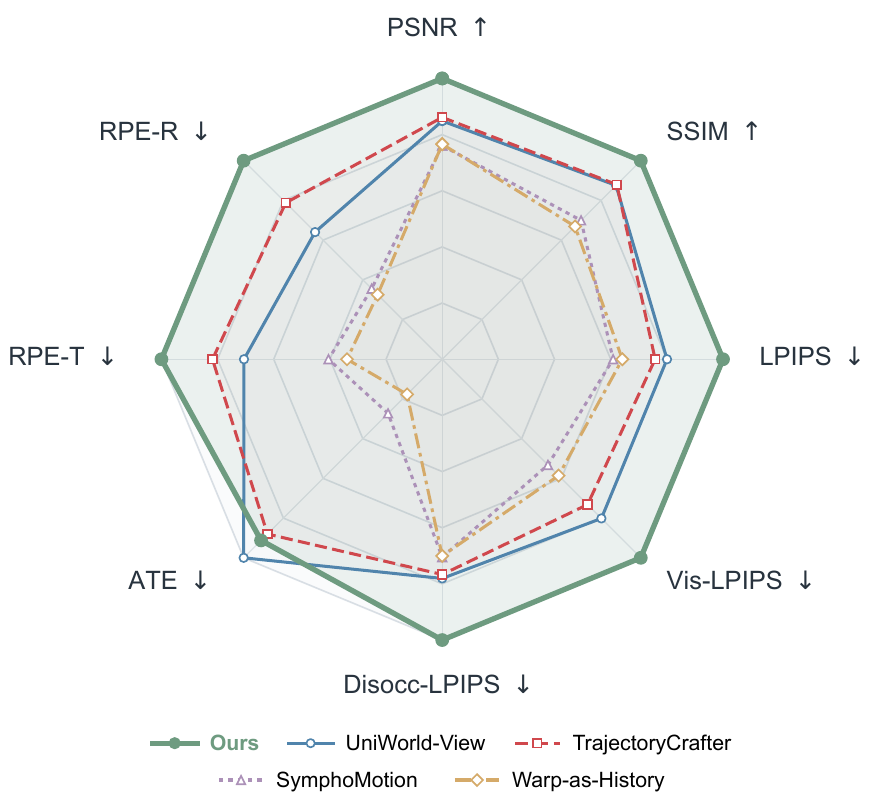}
    \caption{Comparison with existing methods. With limited training data and iterations, our method achieves state-of-the-art visual quality and trajectory accuracy in camera-controlled video generation.}
    \label{fig:teaser}
\end{figure}

\IEEEPARstart{R}{ecent} video diffusion models (VDMs) can synthesize videos with coherent motion, consistent scene content, and plausible viewpoint changes over time~\cite{SVD,Wan21,CogVideoX}. Trained on large-scale video data, these models acquire strong spatiotemporal priors, making them powerful backbones for spatial-intelligence tasks such as 3D/4D generation, video re-rendering, and world modeling~\cite{ViewCrafter,TrajCrafter,NVSSolver,VideoScene,Free4D,Genie,NavigationWM,4DSpatialSurvey}. In these applications, the camera trajectory determines how a scene is observed over time, which regions become visible, and how viewpoint changes relate to scene geometry. Effective trajectory control is therefore essential. The generated video should follow a specified camera path while preserving scene content, object motion, temporal coherence, and visual quality.

To realize such control, existing approaches broadly follow two directions. One direction adapts VDMs for camera control through additional trainable modules, pose-aware conditioning interfaces, or model fine-tuning~\cite{MotionCtrl,CameraCtrl,ReCamMaster,TrajectoryAttention,TrajCrafter,GoWithTheFlow}. These methods provide explicit camera control but often require large-scale trajectory--video supervision or careful fine-tuning. With limited supervision, such adaptation may weaken the visual and temporal priors of the pretrained model. Recent work has therefore explored training-free alternatives that keep the pretrained VDM fixed and inject trajectory cues only during inference~\cite{NVSSolver,ViewExtrapolator}, including the zero-shot variant of Warp-as-History~\cite{WarpAsHistory}. A representative route is \emph{warping-and-repainting}, which warps already observed content along the target camera path and uses the pretrained video model to synthesize newly revealed regions. This route enables direct camera control at inference time while preserving the pretrained model, making it well suited to downstream spatial-intelligence tasks.

Despite these advantages, warping-and-repainting can introduce a denoising-state mismatch during diffusion sampling. When warped visible regions are directly combined with model-generated regions in a noisy latent, the two parts may correspond to different noise states. The warped regions are associated with externally sampled noise, whereas the generated regions follow the model's own denoising trajectory. We refer to this mismatch as a \emph{denoising-state domain gap}. It can lead to boundary discontinuities, geometric artifacts, noisy patches, and texture degradation. The key challenge is therefore to inject useful trajectory evidence from warped references while reducing the denoising-state gap introduced by their fusion.

To address this problem, we propose \textbf{WorldForge V2}, a camera-trajectory-control framework that helps VDMs use imperfect warped trajectory signals more effectively while preserving generation quality.
WorldForge V2 follows the warping-and-repainting route.
First, it addresses the incompatibility introduced by warped-reference fusion: direct fusion in noisy latents mixes externally noised warped regions with generated regions following the model's current denoising state.
WorldForge V2 reduces this mismatch by fusing the warped reference into the model-predicted clean estimate and mapping the fused result back to the current denoising state, enabling training-free trajectory-aligned video generation.
Second, clean-estimate fusion provides a reliable way to inject warped references into the denoising process by reducing noisy-state mismatch.
To further improve its sampling efficiency, we introduce \vcache{}, which approximates the clean-estimate correction directly in noisy latents.
In this way, warped-reference guidance can retain the benefit of clean-estimate fusion while reducing the extra sampling overhead.
Finally, full-model fine-tuning and large trainable control modules involve optimizing a large number of parameters, making it difficult to use limited training data efficiently. To address this problem, we build on our pipeline and further investigate an optimization-based approach to camera-trajectory control. Since different layers in VDMs specialize in different functions, we extend the motion-channel localization used in FLF to model adaptation. We identify trajectory-sensitive blocks and apply LoRA only to these blocks. This selective adaptation provides a data-efficient alternative to broad model tuning.

First, we address the denoising-state incompatibility caused by warped-reference fusion. In direct noisy-space fusion, the warped visible regions and the model-generated regions may follow different denoising states, causing geometric artifacts, boundary discontinuities, and noisy patches. To mitigate this mismatch, Intra-Step Recursive Refinement (IRR) fuses the warped reference with the model-predicted clean estimate under the visibility mask and then re-noises the fused estimate to the current timestep. This fuse-then-renoise process associates all regions with the same Gaussian noise realization, thereby reducing their denoising-state gap.
Camera-trajectory control also faces a typical motion--appearance disentanglement problem, as enforcing camera motion may inadvertently alter object and scene appearance. To address this problem, Flow-Gated Latent Fusion (FLF) uses optical-flow cues to locate channels that are highly correlated with motion and injects trajectory guidance only into these channels, thereby reducing appearance distortion.
Finally, artifacts or misalignment in the warped reference may degrade the quality of the generated video. To mitigate this issue, we propose Dual-Path Self-Corrective Guidance (DSG). Inspired by classifier-free guidance~\cite{CFG}, DSG compares the trajectory-guided and unguided denoising directions. Their difference captures the update introduced by trajectory guidance, which DSG uses to suppress unreliable deviations while retaining useful trajectory constraints. Together, IRR, FLF, and DSG enable training-free trajectory-aligned video generation while reducing the adverse effects of incompatible denoising states and unreliable reference signals.

Second, the fuse-then-renoise process in IRR provides an effective way to reduce the denoising-state gap introduced by warped-reference fusion. We further improve the efficiency of this process with \vcache{}. Our key observation is that the correction induced by clean-estimate fusion can be approximated directly in noisy latent space from a denoising prediction already available during sampling. Based on this observation, \vcache{} applies the approximated correction during warped-reference fusion, allowing trajectory guidance to remain on the original denoising path. This avoids an additional model evaluation and reduces inference latency.

Finally, we further develop an optimization-based extension of our pipeline for camera-trajectory control. Existing optimization-based methods often fine-tune the full model or introduce large trainable control modules, making inefficient use of limited supervision. Since camera control mainly depends on motion-related representations, we extend the motion-channel localization idea of FLF to model adaptation. Leveraging the functional specialization of different VDM layers~\cite{MOFT}, we identify trajectory-sensitive blocks, apply LoRA only to these blocks, and train the adapters with a two-stage optimization strategy tailored to our pipeline. This selective adaptation provides a data-efficient alternative to broad model tuning. In our experiments, we find that training data from a single scene is sufficient to enable effective trajectory following.

Our main contributions are summarized as follows:

\begin{itemize}

    \item To address denoising-state mismatch, motion--appearance entanglement, and unreliable warped references, we develop a training-free guidance pipeline with IRR, FLF, and DSG, enabling trajectory-aligned video generation with fewer artifacts and less appearance distortion.

    \item To retain the benefit of clean-estimate correction with lower sampling cost, we introduce \vcache{}, which approximates the correction in noisy latent space and avoids an additional model evaluation, thereby reducing inference latency.

    \item To avoid data-inefficient full-model tuning, we identify trajectory-sensitive blocks, adapt only these blocks with LoRA, and optimize them using a two-stage strategy tailored to our pipeline; training data from a single scene is sufficient to enable effective trajectory following.

    \item Comparative experiments on multiple public datasets demonstrate accurate camera-path control and high-quality video generation.
\end{itemize}

A preliminary version of this work was published at \textbf{CVPR 2026~\cite{worldforge}}. The source code is available at \url{https://github.com/Westlake-AGI-Lab/WorldForge}.

\begin{figure*}[!t]
    \centering
    \includegraphics[width=0.95\linewidth]{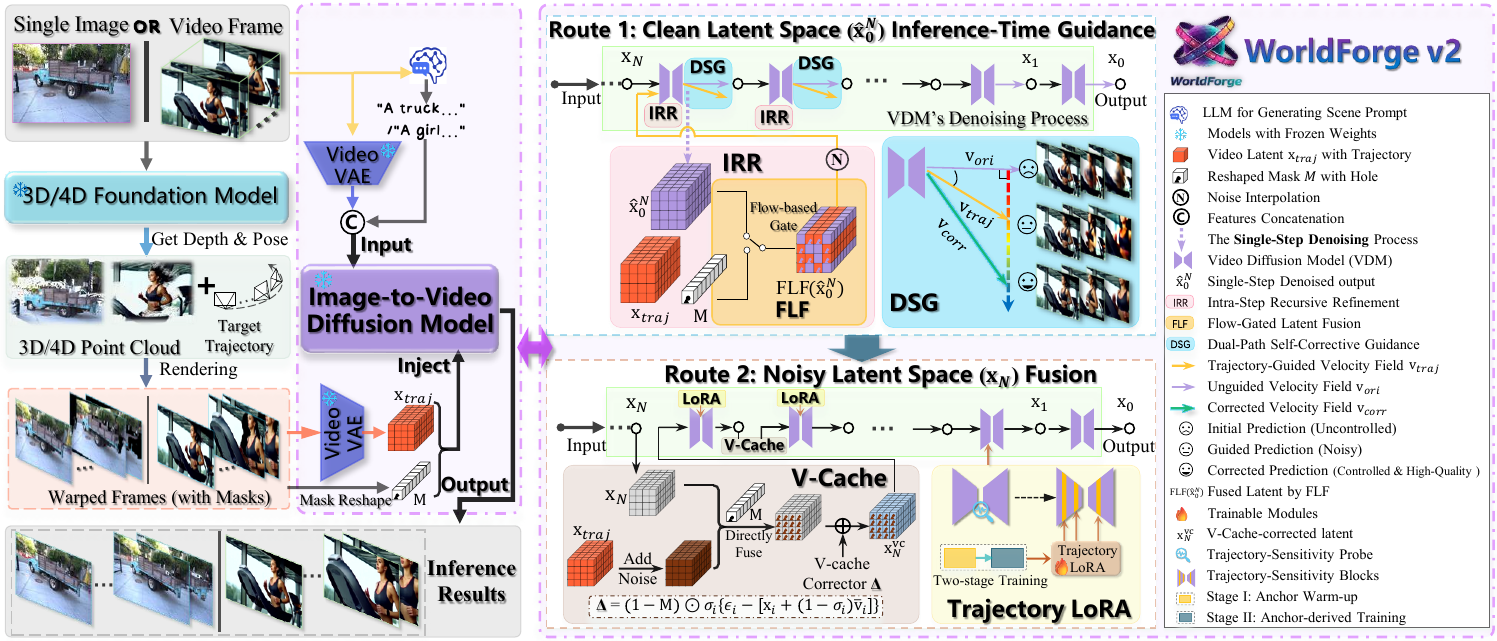}\caption{
        \textbf{Overview of WorldForge V2.}
        Given a single image or video frame and a target camera trajectory, the scene geometry is reconstructed and rendered into trajectory-aligned warped frames.
        WorldForge V2 first performs training-free trajectory guidance in the predicted clean latent space using IRR, FLF, and DSG.
        Building on this framework, we further introduce an efficient route that directly fuses trajectory information in the noisy latent space with V-Cache, together with a trajectory-sensitive lightweight LoRA scheme for camera trajectory control using only limited training data.
        The resulting video follows the specified camera path while preserving the generative capability of the pretrained video diffusion model.
        }
    \label{fig:pipeline}
    \end{figure*}

\section{Related Work}
\label{sec}

We review four lines of work closely related to WorldForge V2: video diffusion priors for spatial generation, camera control in video diffusion models, guidance and correction strategies in generative models, and parameter-efficient adaptation for camera control.

\subsection{Video Diffusion Priors for Spatial Generation}

Video diffusion models (VDMs) learn rich priors of real-world appearance, motion, temporal continuity, and viewpoint change from large-scale video data~\cite{SVD,Wan21,CogVideoX,HunyuanVideo}. Recent studies have transferred these priors to spatial generation tasks, including 3D/4D scene generation, novel-view video synthesis, and world modeling~\cite{AC3D,VD3D,ViewCrafter,TrajCrafter,NVSSolver,ReCamMaster,GCD,Free4D,NavigationWM,Genie,LingBotWorld,InSpatioWorld,HYWorld2,WorldPlay,MatrixGame3}. This scene-level use of video priors complements explicit scene reconstruction based on NeRFs, Gaussian splatting, multi-view and light-field synthesis~\cite{NeRF,3DGS,MVSNet,Mipsplatting,InstantNGP,Fewarnet,PDFSV}, as well as scene generation from panoramic images~\cite{PERF} and object-centric 3D generation based on diffusion models~\cite{DreamFusion,ASD,Trellis,GeometryAwareSDS,DreamComposerPlusPlus}. While reconstruction methods recover scene structure from observed views, VDMs can use their learned priors to generate plausible content beyond the observed regions. However, VDM pretraining does not explicitly align camera trajectories with scene content. As a result, these models can generate high-quality videos but cannot reliably follow a prescribed camera path, motivating growing research on trajectory-controlled video generation. WorldForge V2 combines warped geometric evidence with the generative prior of a pretrained VDM to control viewpoint changes.

\subsection{Camera Control in Video Diffusion Models}

Existing camera-control methods broadly follow two directions. Training-based methods introduce pose-aware conditioning interfaces, additional control modules, or model fine-tuning to learn camera-controlled generation~\cite{MotionCtrl,CameraCtrl,TrajectoryAttention,ReCamMaster,TrajCrafter,SyncamMaster,GoWithTheFlow,FYC}. These methods provide explicit camera control but usually require substantial trajectory--video supervision; when training data are limited, model adaptation may weaken the visual and temporal priors learned during pretraining. Recent work has therefore increasingly explored training-free inference-time control and lightweight adaptation, which preserve most or all of the pretrained model and inject trajectory cues through initial noise, latent transformations, rendered references, or history frames~\cite{Free4D,NVSSolver,ViewExtrapolator,VUD,LatentReframe,WarpAsHistory,WorldInWorld}. Across training-based and training-free methods, warping-and-repainting warps observed content along the target camera path and uses a VDM to generate newly revealed regions~\cite{See3D,Voyager}. This route provides explicit geometric correspondence and integrates naturally with depth estimation, point-based rendering, and visibility masks. WorldForge V2 adopts this route and studies how warped references can be used more effectively for camera-trajectory control while reducing the artifacts introduced during denoising. Related motion-control work also uses entity-specific 3D trajectories to control object motion~\cite{TrajMaster3D}.

\subsection{Guidance and Correction Strategies in Generative Models}
Guidance methods steer diffusion or flow models toward desired outputs by modifying their sampling directions. Classifier-free guidance (CFG) combines conditional and unconditional predictions to improve alignment with semantic conditions~\cite{CFG}. Subsequent methods further adjust the guidance direction through negative or auxiliary branches, differential predictions, adaptive rescaling, resampling, and iterative repainting~\cite{BadGuide,STG,APG,FlowDirector,RestartSampling,Repaint,zigzag}. In 3D/4D generation, guidance is also widely used to enforce viewpoint consistency from rendered or warped observations~\cite{D2T,VistaDream,NVSSolver,ViewExtrapolator,Free4D}. However, conventional guidance does not directly address the artifacts introduced by noisy warped references. WorldForge V2 explores correction strategies for warping-guided generation, including correcting warped-reference fusion in noisy latent space and using the difference between guided and unguided predictions to refine the denoising process. Together, these strategies improve the quality of trajectory-controlled videos generated under warping guidance.

\subsection{Parameter-Efficient Adaptation for Camera Control}

Model adaptation commonly uses trainable control branches or parameter-efficient fine-tuning. ControlNet and T2I-Adapter introduce lightweight conditioning modules, while LoRA adapts pretrained models through low-rank parameter updates~\cite{ControlNet,T2IAdapter,LoRA}. These lightweight adaptation strategies have also been applied to motion learning, sparse-view synthesis, and trajectory-controlled generation with limited training data~\cite{CustomizeAVideo,FrameCrafter,WarpAsHistory,FYC}. Related studies further show that motion- and camera-related representations are concentrated in particular latent channels or network layers~\cite{MOFT,LiONLoRA}. Based on these findings, WorldForge V2 uses optical-flow cues in FLF to localize motion-related latent channels and reduce appearance interference. We further extend this idea to model adaptation by identifying trajectory-sensitive blocks and applying lightweight LoRA only to these blocks, enabling effective trajectory alignment with very limited training data.

\section{Method}
\label{sec:method}

WorldForge V2 (Fig.~\ref{fig:pipeline}) aims to control camera trajectories while preserving the video priors of pretrained generation models. Given a source image or video and a target camera path, geometric warping first produces a trajectory reference and its visibility mask. WorldForge V2 then uses these signals to guide denoising and generate videos that follow the target path. Sec.~\ref{sec:preliminaries} introduces the required notation and background. Sec.~\ref{sec:trajectory_guidance} presents the training-free guidance pipeline: IRR reduces the denoising-state gap through clean-estimate fusion, FLF restricts trajectory injection to motion-related latent channels, and DSG corrects the guided prediction using the model's original denoising path. Sec.~\ref{sec:vcache} derives a noisy-latent correction that reduces the denoising-state gap with fewer model evaluations. Finally, Sec.~\ref{sec:lora} identifies trajectory-sensitive network blocks and adapts them with LoRA using limited supervision, enabling data-efficient trajectory control for VDMs.

\subsection{Preliminaries}
\label{sec:preliminaries}

We introduce the notation and background required by WorldForge V2, including flow-based denoising, sampling-time guidance, low-rank adaptation, and geometry-based trajectory guidance.

\noindent\textbf{Flow-Based Denoising.}
Diffusion and flow-based models both generate videos by iteratively transforming a noisy latent into a clean sample. A diffusion process has a corresponding probability-flow ODE and can be written in flow form through reparameterization~\cite{DDPM,Flow,PFODE,DiffusionVisionSurvey}; the detailed relationship is provided in the supplementary material.

Our implementation uses flow-based video models~\cite{Wan21,Longcat}, so we describe sampling in velocity form. Let $\mathbf{x}_i$ be the noisy video latent at noise level $\sigma_i$, and let $\ell$ denote the model condition. The pretrained model $f_{\theta}$ predicts
$\mathbf{v}_i=f_{\theta}(\mathbf{x}_i,\sigma_i,\ell)$.
A first-order solver updates the latent as
\begin{equation}
\label{eq:base_update}
\mathbf{x}_{i-1}
=
\mathbf{x}_i
+
(\sigma_{i-1}-\sigma_i)\mathbf{v}_i,
\qquad
\sigma_{i-1}<\sigma_i.
\end{equation}

The same velocity also gives the one-step clean estimate
\begin{equation}
\label{eq:flow_estimates}
\hat{\mathbf{x}}_{0,i}
=
\mathbf{x}_i-\sigma_i\mathbf{v}_i.
\end{equation}

Thus, $\mathbf{x}_i$ is the current noisy state, whereas $\hat{\mathbf{x}}_{0,i}$ is the clean content predicted from it. WorldForge V2 injects trajectory evidence by modifying either of these states before the next solver update.

\noindent\textbf{Classifier-Free Guidance.}
Classifier-free guidance (CFG)~\cite{CFG} strengthens conditioning by comparing conditional and unconditional predictions. In the velocity formulation,
\begin{equation}
\label{eq:cfg_velocity}
\tilde{\mathbf{v}}_i
=
f_{\theta}
\left(
\mathbf{x}_i,\sigma_i,\ell
\right)
+
\omega_{\mathrm{CFG}}
\left[
f_{\theta}
\left(
\mathbf{x}_i,\sigma_i,\ell
\right)
-
f_{\theta}
\left(
\mathbf{x}_i,\sigma_i,\emptyset
\right)
\right],
\end{equation}
where $\omega_{\mathrm{CFG}}$ is the guidance weight and $\emptyset$ denotes the unconditional input. The prediction difference measures the update introduced by the condition. DSG later follows this principle by comparing denoising predictions with and without trajectory guidance.

\noindent\textbf{Low-Rank Adaptation.}
Low-Rank Adaptation (LoRA)~\cite{LoRA} keeps a pretrained weight $\mathbf{W}\in\mathbb{R}^{d_{\mathrm{out}}\times d_{\mathrm{in}}}$ fixed and represents its trainable update with two low-rank matrices:
\begin{equation}
\label{eq:lora_prelim}
\mathbf{W}'
=
\mathbf{W}
+
\lambda_{\mathrm{L}}\mathbf{B}\mathbf{A},
\end{equation}
where $\mathbf{A}\in\mathbb{R}^{r\times d_{\mathrm{in}}}$ and $\mathbf{B}\in\mathbb{R}^{d_{\mathrm{out}}\times r}$ are trainable, $r\ll\min(d_{\mathrm{in}},d_{\mathrm{out}})$, and $\lambda_{\mathrm{L}}$ controls the adapter scale. LoRA therefore adapts a model with far fewer trainable parameters than full-model fine-tuning. Sec.~\ref{sec:lora} applies LoRA only to trajectory-sensitive network blocks.

\noindent\textbf{Geometry-Based Warping and Trajectory Guidance.}
WorldForge V2 converts a target camera path into explicit trajectory evidence through geometric warping. Given a source image or video, we obtain depth and camera poses using available geometry estimators, such as VGGT and UniDepth~\cite{VGGT,UniDepth}. For source frame $\mathbf{I}_{\mathrm{src}}$, source depth $\mathbf{D}_{\mathrm{src}}$, source pose $\mathbf{P}_{\mathrm{src}}$, and target pose $\mathbf{P}_{\mathrm{tar}}$, the warping operator $\mathcal{W}$ produces a partial target view and its visibility mask:
\begin{equation}
\label{eq:warp}
(\mathbf{I}'_{\mathrm{tar}},\mathbf{M}_{\mathrm{tar}})
=
\mathcal{W}(\mathbf{I}_{\mathrm{src}},\mathbf{D}_{\mathrm{src}},
\mathbf{P}_{\mathrm{src}},\mathbf{P}_{\mathrm{tar}}).
\end{equation}

We denote the VAE-encoded warped reference by $\mathbf{x}_{\mathrm{traj}}$ and its latent-space mask by $\mathbf{M}$. This partial reference follows the target path but contains holes, disocclusions, and depth-boundary errors.

A common way to use this reference is to noise it to the current level and insert it into the visible region of $\mathbf{x}_i$. Let $\boldsymbol{\epsilon}_i$ be the Gaussian noise used at level $\sigma_i$. Direct noisy-latent insertion is
\begin{equation}
\label{eq:direct_noisy_insertion}
\mathbf{x}^{\mathrm{dir}}_i =
(1-\mathbf{M})\odot\mathbf{x}_i
+ \mathbf{M}\odot
\left[(1-\sigma_i)\mathbf{x}_{\mathrm{traj}}
+ \sigma_i\boldsymbol{\epsilon}_i\right].
\end{equation}

This operation combines an externally re-noised visible region with a model-generated region that follows the current denoising path. Their different noise states may create a denoising-state gap. Sec.~\ref{sec:trajectory_guidance} reduces this gap through clean-estimate fusion, while Sec.~\ref{sec:vcache} derives a more efficient noisy-latent approximation.

Camera trajectories can also be injected through other interfaces. Warp-as-History provides warped frames through the model's history interface~\cite{WarpAsHistory}, whereas Go-with-the-Flow injects warped trajectory signals into the initial noise or initial denoising state~\cite{GoWithTheFlow}. We later use these interfaces to evaluate whether the trained LoRA transfers to different trajectory signals without additional fine-tuning.

\subsection{Inference-Time Trajectory Guidance}
\label{sec:trajectory_guidance}

Geometric warping explicitly indicates how observed content should move along the target camera path. However, directly inserting the noised reference mixes the externally re-noised visible region with the model-generated region, producing a denoising-state gap that can cause noisy patches and structural errors. We therefore inject the reference through the model-predicted clean estimate and re-noise the fused result with a unified noise sample. We further restrict guidance to motion-related channels and calibrate the guided prediction against the model's original denoising path. Together, these designs enable training-free trajectory control while preserving visual quality.

\noindent\textbf{Intra-Step Recursive Refinement (IRR).}
Direct noisy-latent insertion assigns newly sampled noise to the warped visible region while retaining the model's current denoising state elsewhere. To avoid directly combining these incompatible states, IRR first fuses the warped reference in the model-predicted clean space and then re-noises the fused result with one Gaussian noise sample.

At denoising step $i$, IRR obtains the one-step clean estimate $\hat{\mathbf{x}}_{0,i}$ from Eq.~\eqref{eq:flow_estimates} and fuses the warped reference under the visibility mask:
\begin{equation}
\label{eq:irr_clean_fusion}
\tilde{\mathbf{x}}_{0,i}
=
\mathbf{M}\odot\mathbf{x}_{\mathrm{traj}}
+
(1-\mathbf{M})\odot\hat{\mathbf{x}}_{0,i}.
\end{equation}

The visible region follows the warped reference, while the remaining region retains the model-predicted clean content. IRR then samples $\boldsymbol{\epsilon}_i$, maps the fused estimate back to the current noise level, and predicts the corresponding velocity:
\begin{equation}
\label{eq:irr_state}
\begin{aligned}
\mathbf{x}^{\mathrm{irr}}_i
&=
(1-\sigma_i)\tilde{\mathbf{x}}_{0,i}
+
\sigma_i\boldsymbol{\epsilon}_i,
\\
\mathbf{v}^{\mathrm{irr}}_i
&=
f_{\theta}
\left(
\mathbf{x}^{\mathrm{irr}}_i,
\sigma_i,
\ell
\right).
\end{aligned}
\end{equation}

Because all regions of $\mathbf{x}^{\mathrm{irr}}_i$ are re-noised with the same $\boldsymbol{\epsilon}_i$, the warped and generated regions no longer come from directly stitched noisy states. IRR therefore preserves the trajectory evidence while reducing their denoising-state gap.

\noindent\textbf{Flow-Gated Latent Fusion (FLF).}
Camera-trajectory control should separate camera-induced motion from object appearance and intrinsic motion. However, VAE channels encode different types of information, so replacing all channels with the same trajectory signal may also distort appearance. This channel specialization is supported by MOFT, which uses PCA to identify motion-related channels~\cite{MOFT}. Based on this observation, FLF uses optical flow to select channels whose temporal motion agrees with the target trajectory, requiring no gradient-based optimization during inference.

For denoising step $i$ and channel $c$, FLF estimates optical flow between adjacent frames of $\hat{\mathbf{x}}_{0,i}^{\,c}$ and $\mathbf{x}_{\mathrm{traj}}^{\,c}$. We use Farneback by default and evaluate RAFT as an alternative~\cite{OpticaFlow,RAFT}. Within the visible region, normalized endpoint error $E_i^c$, angular error $A_i^c$, and outlier ratio $O_i^c$ measure displacement, direction, and reliability differences. We combine them as
\begin{equation}
\label{eq:flf_score}
S_i^c
=
\gamma_E(1-E_i^c)
+
\gamma_A(1-A_i^c)
+
\gamma_O(1-O_i^c),
\end{equation}
where a higher $S_i^c$ indicates stronger agreement with the warped trajectory.

FLF selects $\mathcal{C}_i=\{c\mid S_i^c\geq\delta_i\}$ with adaptive threshold $\delta_i=\mu_i^S-\lambda_i \sigma_i^S$, where $\mu_i^S$ and $\sigma_i^S$ are the mean and standard deviation of channel scores. The schedule keeps more channels during early noisy steps to establish coarse motion and gradually tightens the selection as appearance details emerge. Detailed normalization, weights, and schedules are provided in the supplementary material.

FLF replaces full-channel fusion with
\begin{equation}
\label{eq:flf_fusion}
\tilde{\mathbf{x}}_{0,i}^{\,c}
=
\begin{cases}
\mathbf{M}\odot\mathbf{x}_{\mathrm{traj}}^{\,c}
+
(1-\mathbf{M})\odot
\hat{\mathbf{x}}_{0,i}^{\,c},
&
c\in\mathcal{C}_i,
\\[1mm]
\hat{\mathbf{x}}_{0,i}^{\,c},
&
c\notin\mathcal{C}_i.
\end{cases}
\end{equation}

The selected estimate $\tilde{\mathbf{x}}_{0,i}^{\,c}$ is substituted for $\tilde{\mathbf{x}}_{0,i}$ in Eq.~\eqref{eq:irr_state} and then re-noised.
Thus, trajectory evidence is injected into channels with clear motion agreement, while the remaining channels continue to follow the pretrained model, reducing texture and appearance distortion.

\noindent\textbf{Dual-Path Self-Corrective Guidance (DSG).}
Warped trajectory references may contain distortions caused by depth errors, occlusions, or misalignment, and these errors can degrade the generated video. Inspired by CFG and AutoGuidance~\cite{CFG,BadGuide}, DSG compares a trajectory-guided prediction with the model's original prediction and uses their difference to refine the guidance, allowing the result to follow the trajectory constraint while improving visual quality.

At step $i$, the base velocity $\mathbf{v}_i=f_{\theta}(\mathbf{x}_i,\sigma_i,\ell)$ follows the pretrained denoising path and preserves visual fidelity, but does not enforce the target camera path. The guided velocity $\mathbf{v}^{\mathrm{irr}}_i$ follows the trajectory but may inherit errors from the warped reference. DSG therefore uses $\mathbf{v}_i$ as a quality reference to correct $\mathbf{v}^{\mathrm{irr}}_i$.

We empirically find that the cosine similarity between the two velocities is typically $0.3$--$0.6$, corresponding to an angle of about $50^{\circ}$--$70^{\circ}$, whereas the two branches in standard CFG are nearly parallel. This larger difference arises because IRR and FLF modify the latent itself rather than only the condition. Direct CFG extrapolation can therefore produce excessive updates. DSG instead matches the two velocity magnitudes and uses the trajectory-guided component orthogonal to the base direction, following the principle of APG~\cite{APG}:
\begin{equation}
\label{eq:dsg_velocity}
\mathbf{v}^{\mathrm{corr}}_i
=
\mathbf{v}^{\mathrm{irr}}_i
+
\rho\,\beta_i
\left[
\mathbf{v}^{\mathrm{irr}}_i
-
\alpha_i
\frac{
\left\|\mathbf{v}^{\mathrm{irr}}_i\right\|_2
}{
\left\|\mathbf{v}_i\right\|_2+\varepsilon
}
\mathbf{v}_i
\right].
\end{equation}

Here, $\alpha_i=\langle\mathbf{v}^{\mathrm{irr}}_i,\mathbf{v}_i\rangle/(\|\mathbf{v}^{\mathrm{irr}}_i\|_2\|\mathbf{v}_i\|_2)$ measures directional agreement, $\beta_i=\sqrt{1-\alpha_i^2}$ reflects angular difference, and $\rho$ controls the correction strength. When the two paths agree, the correction remains small; when they diverge, the trajectory-specific component receives more weight. The corrected velocity $\mathbf{v}^{\mathrm{corr}}_i$ is then used for the solver update, preserving useful camera control while reducing artifacts from inaccurate warped references.

Overall, IRR injects the warped reference through clean-estimate fusion, FLF restricts guidance to motion-related channels, and DSG calibrates the guided prediction using the base denoising path. Together, they enable training-free camera-trajectory control while preserving scene appearance and reducing artifacts from imperfect warped references.

\subsection{Efficient Noisy-Latent Correction}
\label{sec:vcache}

The fuse-then-renoise process in IRR effectively reduces the denoising-state gap introduced by warped-reference fusion. Building on this, we further improve its sampling efficiency. IRR first evaluates the video model on the original latent to obtain the clean estimate and then evaluates the trajectory-guided latent to continue denoising. Our key observation is that the correction introduced by clean-estimate fusion can be written directly in noisy latent space. We therefore derive this correction and approximate it using the velocity cached from the preceding denoising step. This gives \vcache{}, a single-forward path for efficient trajectory guidance.

\noindent\textbf{Noisy-Latent Correction.}
To identify the correction required by direct noisy-latent fusion, we compare the IRR state and direct-insertion state constructed from the same $\mathbf{x}_i$, $\mathbf{x}_{\mathrm{traj}}$, $\mathbf{M}$, and $\boldsymbol{\epsilon}_i$. The IRR state first forms $\hat{\mathbf{x}}_{0,i}=\mathbf{x}_i-\sigma_i\mathbf{v}_i$, fuses it with the warped reference, and re-noises the result. The direct-insertion state instead inserts the noised reference in the visible region and retains $\mathbf{x}_i$ elsewhere. Their difference is
\begin{equation}
\label{eq:vcache_difference}
\begin{aligned}
&\mathbf{x}^{\mathrm{irr}}_i-\mathbf{x}^{\mathrm{dir}}_i \\
&=
(1-\sigma_i)
\Bigl[
\mathbf{M}\odot\mathbf{x}_{\mathrm{traj}}
+(1-\mathbf{M})\odot(\mathbf{x}_i-\sigma_i\mathbf{v}_i)
\Bigr]
+\sigma_i\boldsymbol{\epsilon}_i
\\
&\quad-
\Bigl\{
(1-\mathbf{M})\odot\mathbf{x}_i
+\mathbf{M}\odot
\bigl[
(1-\sigma_i)\mathbf{x}_{\mathrm{traj}}
+\sigma_i\boldsymbol{\epsilon}_i
\bigr]
\Bigr\}
\\
&=
(1-\mathbf{M})\odot
\Bigl[
(1-\sigma_i)(\mathbf{x}_i-\sigma_i\mathbf{v}_i)
+\sigma_i\boldsymbol{\epsilon}_i
-\mathbf{x}_i
\Bigr]
\\
&=
(1-\mathbf{M})\odot\sigma_i
\bigl\{
\boldsymbol{\epsilon}_i
-
\underbrace{
\bigl[\mathbf{x}_i+(1-\sigma_i)\mathbf{v}_i\bigr]
}_{\substack{\text{\scriptsize one-step inversion noise}}}
\bigr\}.
\end{aligned}
\end{equation}

The visible region cancels because both states use the same re-noised warped reference there. The remaining difference lies in the model-generated region. Under the flow formulation, $\mathbf{x}_i+(1-\sigma_i)\mathbf{v}_i$ is the one-step inversion noise inferred from the current latent and velocity. Eq.~\eqref{eq:vcache_difference} therefore shows that IRR differs from direct insertion by a model-dependent noise correction outside the visible mask, making the two regions more compatible at the current noise level.

\noindent\textbf{Cached Approximation.}
The correction in Eq.~\eqref{eq:vcache_difference} requires the current base velocity $\mathbf{v}_i$, adding a model evaluation on $\mathbf{x}_i$. Diffusion caching methods reuse intermediate computations by exploiting the correlation between predictions at nearby denoising steps~\cite{TeaCache}. Following this idea, \vcache{} approximates $\mathbf{v}_i$ with the velocity cached from the preceding step, denoted by $\bar{\mathbf{v}}_i$. Starting from the second guided step, the corrected latent is
\begin{equation}
\label{eq:vcache_state}
\mathbf{x}^{\mathrm{vc}}_i
=
\mathbf{x}^{\mathrm{dir}}_i
+
(1-\mathbf{M})\odot\sigma_i
\left[
\boldsymbol{\epsilon}_i
-
\left(
\mathbf{x}_i
+
(1-\sigma_i)\bar{\mathbf{v}}_i
\right)
\right].
\end{equation}

The first guided step has no cached velocity and therefore uses direct noisy-latent insertion. At subsequent steps, the model is evaluated once on $\mathbf{x}^{\mathrm{vc}}_i$ to obtain $\mathbf{v}^{\mathrm{vc}}_i=f_{\theta}(\mathbf{x}^{\mathrm{vc}}_i,\sigma_i,\ell)$; this velocity updates the sampler and is cached for the next step. Thus, \vcache{} approximates the IRR correction with one model evaluation per guided step. The experiments compare this approximation with the exact current-step correction in terms of noise statistics and generation quality.

\subsection{Trajectory-Sensitive LoRA Adaptation}
\label{sec:lora}

Optimization-based camera control often relies on full-model fine-tuning or large trainable modules, requiring substantial trajectory--video data and computation. With limited supervision, broad parameter updates may also weaken pretrained visual and spatiotemporal priors, causing degraded geometry or implausible structures. We therefore seek a data-efficient adaptation strategy that strengthens trajectory control with limited supervision while restricting parameter updates to preserve the pretrained priors.

Our method combines trajectory-sensitive block selection with \vcache{}-aligned LoRA training. FLF analyzes how motion and appearance features are distributed across latent channels; we extend this localization idea to network blocks and apply LoRA only to blocks that are sensitive to trajectory signals. We then inject trajectories through \vcache{} and train these adapters on latent states produced by \vcache{} guidance. This design concentrates limited supervision on trajectory-relevant components and enables data-efficient adaptation.

\noindent\textbf{Trajectory-Supervised Data Construction.}
To construct paired trajectory references and target videos, we use the public DL3DV dataset~\cite{DL3DV}, which provides multi-view observations and optimized 3D Gaussian Splatting representations~\cite{3DGS}. For each scene, we select multiple source views and sample several target camera paths for each view. The 3DGS representation provides the source-view depth and renders RGB frames along each target path.

Using the source depth, we warp each source image along the sampled path according to Eq.~\eqref{eq:warp}, producing a warped reference and visibility mask. We render the same 3DGS scene along the identical path to obtain the clean target video. Each pair therefore shares the same trajectory: the warped video provides the trajectory condition, and the rendered video provides geometrically aligned supervision. Video-VAE encoding yields the trajectory latent $\mathbf{x}_{\mathrm{traj}}$ and clean target latent $\mathbf{x}^{\mathrm{gt}}_0$.

\noindent\textbf{Trajectory-Sensitive Block Identification.}
Prior analyses suggest that motion-related responses are not uniformly distributed across video-model layers~\cite{MOFT,LiONLoRA}. To identify blocks that are sensitive to trajectory control, we extend the motion--appearance localization idea of FLF from latent channels to network blocks. We treat all self-attention blocks as candidates and determine their sensitivity directly from their responses to trajectory supervision.

To measure this sensitivity, we attach a lightweight probe LoRA to every candidate block while keeping the pretrained model fixed. A small set of warped-reference--target pairs is encoded into latent space. At a sampled noise level $\sigma$, we interpolate between the clean target and trajectory latent and train the probe with
\begin{equation}
\label{eq:probe_training}
\begin{aligned}
\mathbf{x}^{\mathrm{probe}}_{\sigma}
&=
(1-\sigma)\mathbf{x}^{\mathrm{gt}}_0
+
\sigma\mathbf{x}_{\mathrm{traj}},
\\
\mathcal{L}_{\mathrm{probe}}
&=
\mathbb{E}
\left[
\left\|
\mathbf{v}_{\theta,\phi_{\mathrm{p}}}
\left(
\mathbf{x}^{\mathrm{probe}}_{\sigma},
\sigma,
\ell
\right)
-
\frac{
\mathbf{x}^{\mathrm{probe}}_{\sigma}
-
\mathbf{x}^{\mathrm{gt}}_0
}{
\sigma
}
\right\|_2^2
\right],
\end{aligned}
\end{equation}
where $\phi_{\mathrm{p}}$ denotes the probe parameters. The probe learns the denoising direction from the warped trajectory state to the clean target at different noise levels. Brief probe training provides the responses needed to measure block sensitivity.

After probe training, we estimate each block's contribution using Shapley analysis~\cite{Shapley1953}. We score each block subset by the normalized projection of its velocity change onto the full probe LoRA response, as defined in the supplementary material. The Shapley value $\Phi_b$ averages the marginal change in this score when the LoRA in block $b$ is enabled across sampled block combinations. We rank blocks by $\Phi_b$ and retain LoRA only in $\mathcal{B}_{\mathrm{traj}}=\operatorname{TopK}_{b}(\Phi_b)$ for final adaptation.

\begin{figure*}[!t]
    \centering
    \includegraphics[width=0.95\textwidth]{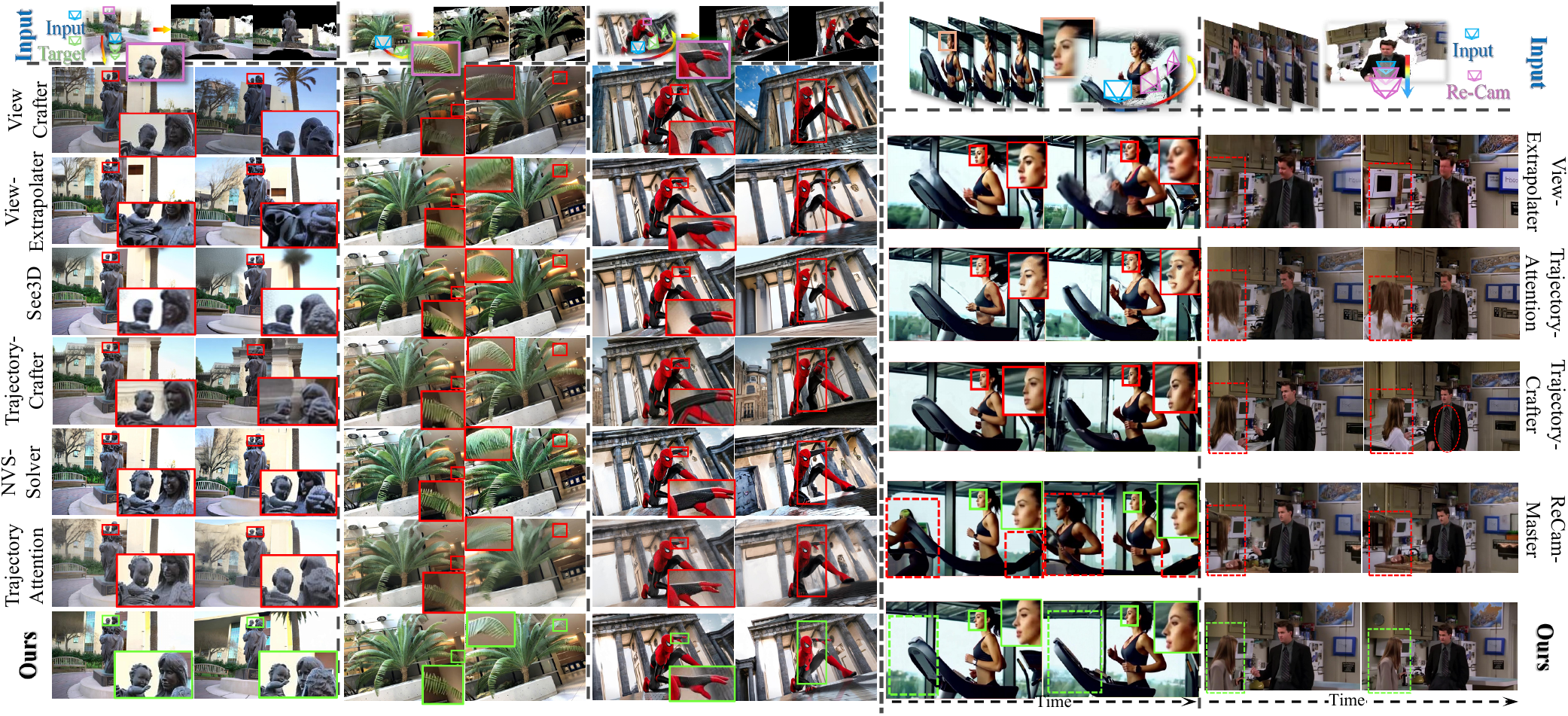}
    \caption{Qualitative comparison for 3D novel-view synthesis and 4D trajectory-controlled re-rendering. For single-image 3D generation (left), our method produces more coherent geometry and plausible newly revealed content. For dynamic re-rendering (right), it preserves human appearance and object motion while reducing floating structures and flattened faces. The training-free guidance yields more realistic results in both static and dynamic scenes.}
    \label{fig:tf_compare}
\end{figure*}

\noindent\textbf{Data-Efficient Trajectory LoRA Training.}
The final LoRA is used with \vcache{}, so its training inputs should reflect the states encountered during \vcache{} inference. Standard flow matching constructs inputs by independently noising the clean target. In contrast, a \vcache{} state also contains the warped reference, visibility mask, and cached inversion-noise correction. Training only on independently noised targets would therefore leave a gap between training and inference states.

To construct \vcache{}-aligned inputs, we use two stages. The first records real intermediate states from \vcache{}-guided denoising and uses them to warm up the selected adapters. The second starts from these states and constructs additional inputs that retain the same warped-reference insertion and \vcache{} correction at sampled noise levels. Both stages optimize
\begin{equation}
\label{eq:lora_training_loss}
\mathcal{L}_{\mathrm{LoRA}}
=
\mathbb{E}
\left[
\left\|
\mathbf{v}_{\theta,\phi}
\left(
\tilde{\mathbf{x}}_{\sigma},
\sigma,
\ell
\right)
-
\frac{
\tilde{\mathbf{x}}_{\sigma}
-
\mathbf{x}^{\mathrm{gt}}_0
}{
\sigma
}
\right\|_2^2
\right],
\end{equation}
where $\tilde{\mathbf{x}}_{\sigma}$ is a \vcache{}-aligned state and $\phi$ denotes the selected LoRA parameters.

In the first stage, we run \vcache{}-guided sampling with a fixed noise schedule and save several intermediate states as anchors. For anchor $\mathbf{x}_a$ at noise level $\sigma_a$, we set $\tilde{\mathbf{x}}_{\sigma_a}=\mathbf{x}_a$. These anchors already contain the warped reference, visibility mask, and cached correction used during inference, allowing LoRA to first adapt to the actual \vcache{} input distribution.

The second stage constructs \vcache{}-consistent inputs at sampled noise levels. Starting from $(\mathbf{x}_a,\sigma_a)$, the warmed-up model predicts $\mathbf{v}_a=\mathbf{v}_{\theta,\phi}(\mathbf{x}_a,\sigma_a,\ell)$ and propagates the anchor to $\sigma_d$:
\begin{equation}
\label{eq:anchor_propagation}
\bar{\mathbf{x}}_d
=
\mathbf{x}_a
-
(\sigma_a-\sigma_d)\mathbf{v}_a.
\end{equation}

We then apply the same warped-reference insertion, visibility mask, and \vcache{} correction used during inference to obtain $\tilde{\mathbf{x}}_{\sigma_d}$. This state preserves the structure of a \vcache{} input while extending training beyond the recorded anchor levels.

The first stage provides real \vcache{} states, while the second constructs aligned states at additional noise levels, reducing the training--inference gap and extending adaptation beyond the saved anchors. Detailed training settings are provided in the supplementary material. At inference, only the LoRA adapters in $\mathcal{B}_{\mathrm{traj}}$ are loaded. Experiments show that training on one scene generalizes to unseen scenes. The trained LongCat adapter also transfers to other trajectory-injection methods without additional fine-tuning.

\section{Experiments}
\label{sec:experiments}
Our experiments examine how rendered-reference guidance combines camera control with the scene priors of pretrained video models. We first evaluate the training-free framework in static 3D generation and dynamic 4D re-rendering, then study V-Cache correction and data-efficient LoRA adaptation. Transfer experiments examine the reuse of the learned control, and ablations establish the roles of guidance, block selection, and training. We denote the IRR--FLF--DSG framework as \emph{Ours (Training-Free)} and the adapted models as \emph{Ours (LongCat)} and \emph{Ours (H3)}.

\subsection{Experimental Setup}
\label{sec:setup}
\paragraph{Implementation}
For training-free camera control, we use Wan2.1 I2V-14B~\cite{Wan21}, with IRR applied during early denoising. We implement Trajectory LoRA on LongCat-Video~\cite{Longcat} and MiniMax-H3 by adapting selected self-attention projections while keeping the pretrained weights fixed. Both models use anchor warm-up followed by training on guided latent states. Detailed training and inference settings are provided in the supplementary material.

\paragraph{Data and evaluation protocol}
The training-free evaluation covers static scenes and dynamic videos from diverse sources. The static 3D setting contains more than 70 source views from over 40 scenes, including LLFF~\cite{LLFF}, Tanks and Temples~\cite{Tanks}, Mip-NeRF 360~\cite{Mipnerf360}, and real or generated images. The dynamic 4D setting contains more than 50 trajectory-controlled outputs from over 30 source videos, including DAVIS~\cite{DAVIS}, movie clips, and videos generated by video diffusion models.

For LoRA adaptation, we construct trajectory--video pairs from optimized 3D Gaussian Splatting representations~\cite{3DGS} of DL3DV scenes~\cite{DL3DV}. Warping a source view along a target camera path provides the partial reference and its visibility mask; rendering the scene along the same path supplies the target video. The main comparison uses 200 test cases, with five random camera trajectories per scene across 40 scenes. Sequences contain 25--93 frames, with the length adapted to each method's supported input format. Training and test scenes are disjoint.

\paragraph{Metrics}
We evaluate visual quality and camera accuracy together. The training-free experiments use FID and CLIP similarity for static scenes, and FVD and CLIP-V similarity for dynamic videos. For Trajectory LoRA, we additionally evaluate target-view reconstruction with pixel-level metrics, PSNR and SSIM, and the perceptual metric LPIPS. Vis-LPIPS and Disocc-LPIPS assess preservation of visible content and completion of newly exposed regions, respectively. ATE measures overall trajectory agreement, while RPE-T and RPE-R measure local translation and rotation errors. Metric implementations are specified in the supplementary material.

\subsection{Inference-Time Camera-Trajectory Control}
\label{sec:main_comparison}
We evaluate whether warped-reference guidance can follow a prescribed camera path while retaining the scene priors of a frozen video backbone. For static 3D generation, we compare with See3D~\cite{See3D}, ViewCrafter~\cite{ViewCrafter}, ViewExtrapolator~\cite{ViewExtrapolator}, TrajectoryAttention~\cite{TrajectoryAttention}, TrajectoryCrafter~\cite{TrajCrafter}, and NVS-Solver~\cite{NVSSolver}. For dynamic 4D re-rendering, we compare with ViewExtrapolator, TrajectoryAttention, and TrajectoryCrafter using matched source videos and target trajectories. ReCamMaster~\cite{ReCamMaster} does not directly accept warped reference videos as trajectory inputs and is therefore included only in the qualitative comparison.

Tables~\ref{tab:tf_static} and~\ref{tab:tf_dynamic} show that Ours (Training-Free) combines accurate camera motion with strong generation quality. It achieves the best results on all static metrics and leads in FVD, CLIP-V similarity, RPE-T, and RPE-R on dynamic videos. Under large viewpoint changes, our method preserves human identity, object motion, and scene structure (Fig.~\ref{fig:tf_compare}). Competing methods more often produce flattened geometry, floating structures, or implausible content in newly revealed regions. The warped references provide explicit camera constraints, while the video model supplies the scene content needed for realistic 3D and 4D generation.

\begin{table}[!t]
\centering
\caption{Quantitative comparison on 3D static scenes using official implementations and identical source inputs. FID and CLIP assess generation quality; ATE, RPE-T, and RPE-R assess trajectory accuracy. Our training-free method leads on all metrics. $\uparrow$: higher is better; $\downarrow$: lower is better. Bold and underlining mark the best and second-best results.}
\label{tab:tf_static}
\footnotesize
\setlength{\tabcolsep}{3pt}
\renewcommand{\arraystretch}{1.05}
\resizebox{\columnwidth}{!}{\begin{tabular}{lrrrrr}
\toprule
Method & FID$\downarrow$ & CLIP$\uparrow$ & ATE$\downarrow$ & RPE-T$\downarrow$ & RPE-R$\downarrow$ \\
\midrule
See3D & 123.26 & \underline{0.941} & 0.091 & \underline{0.089} & \underline{0.250} \\
ViewCrafter & 117.50 & 0.930 & 0.236 & 0.315 & 0.728 \\
ViewExtrapolator & 125.50 & 0.930 & 0.183 & 0.260 & 0.882 \\
TrajectoryAttention & 122.37 & 0.920 & 0.159 & 0.238 & 0.532 \\
TrajectoryCrafter & \underline{111.49} & 0.910 & \underline{0.090} & 0.152 & 0.267 \\
NVS-Solver & 118.64 & 0.937 & 0.224 & 0.268 & 1.056 \\
\midrule
Ours (Training-Free) & \textbf{96.08} & \textbf{0.948} & \textbf{0.077} & \textbf{0.086} & \textbf{0.221} \\
\bottomrule
\end{tabular}}
\end{table}

\begin{table}[!b]
\centering
\caption{Quantitative comparison on 4D dynamic re-rendering using official implementations with matched source videos and target trajectories. FVD and CLIP-V assess video quality; ATE, RPE-T, and RPE-R assess trajectory accuracy. Our method combines the best video quality with the lowest local camera errors. $\uparrow$: higher is better; $\downarrow$: lower is better. Bold and underlining mark the best and second-best results.}
\label{tab:tf_dynamic}
\footnotesize
\setlength{\tabcolsep}{3pt}
\renewcommand{\arraystretch}{1.05}
\resizebox{\columnwidth}{!}{\begin{tabular}{lrrrrr}
\toprule
Method & FVD$\downarrow$ & CLIP-V$\uparrow$ & ATE$\downarrow$ & RPE-T$\downarrow$ & RPE-R$\downarrow$ \\
\midrule
ViewExtrapolator & 108.48 & 0.913 & 1.040 & 1.208 & 4.750 \\
TrajectoryAttention & 106.94 & 0.911 & 0.605 & 1.238 & \underline{3.560} \\
TrajectoryCrafter & \underline{97.31} & \underline{0.923} & \textbf{0.431} & \underline{1.078} & 8.950 \\
\midrule
Ours (Training-Free) & \textbf{93.17} & \textbf{0.938} & \underline{0.527} & \textbf{0.826} & \textbf{2.690} \\
\bottomrule
\end{tabular}}
\end{table}

\begin{table*}[tp]
\centering
\caption{Quantitative comparison of Trajectory LoRA with camera-control baselines. PSNR, SSIM, and LPIPS assess target-view reconstruction; Vis-LPIPS and Disocc-LPIPS assess visible and newly exposed regions. Both adapted backbones improve reconstruction across these metrics, preserving observed content and completing new regions more faithfully. H3 also achieves the lowest local camera errors. Bold and underlining mark the best and second-best results.}
\label{tab:main}
\small
\setlength{\tabcolsep}{3pt}
\renewcommand{\arraystretch}{1.05}
\resizebox{0.8\textwidth}{!}{\begin{tabular}{lrrrrrrrr}
\toprule
Method & PSNR$\uparrow$ & SSIM$\uparrow$ & LPIPS$\downarrow$ & Vis-LPIPS$\downarrow$ & Disocc-LPIPS$\downarrow$ & ATE$\downarrow$ & RPE-T$\downarrow$ & RPE-R$\downarrow$ \\
\midrule
UniWorld-View & 17.15 & 0.5958 & 0.3370 & 0.2799 & 0.4431 & \textbf{0.00625} & 0.2333 & 0.1218 \\
TrajectoryCrafter & 17.41 & 0.5986 & 0.3558 & 0.3066 & 0.4512 & 0.00710 & \underline{0.2013} & \underline{0.0990} \\
SymphoMotion & 15.38 & 0.4760 & 0.4433 & 0.4215 & 0.4908 & 0.02292 & 0.4065 & 0.2196 \\
Warp-as-History & 15.48 & 0.4553 & 0.4206 & 0.3829 & 0.4939 & 0.03533 & 0.4856 & 0.2398 \\
\midrule
Ours (LongCat) & \underline{19.72} & \underline{0.6663} & \underline{0.2748} & \textbf{0.2231} & \underline{0.3640} & 0.00905 & 0.2725 & 0.1324 \\
Ours (H3) & \textbf{20.21} & \textbf{0.6806} & \textbf{0.2698} & \underline{0.2244} & \textbf{0.3459} & \underline{0.00685} & \textbf{0.1647} & \textbf{0.0780} \\
\bottomrule
\end{tabular}}
\end{table*}

\subsection{V-Cache Correction and Data-Efficient LoRA}
\label{sec:efficiency}
The training-free results establish rendered references as an effective source of camera guidance. We next examine two questions: can the clean-estimate correction be approximated directly in noisy latent space to reduce sampling cost, and can limited trajectory supervision be concentrated in the network components most useful for camera control? V-Cache addresses the first question, while Trajectory LoRA uses this selective adaptation to learn control from very little data.

\begin{figure}[!htbp]
\centering
\includegraphics[width=0.95\columnwidth]{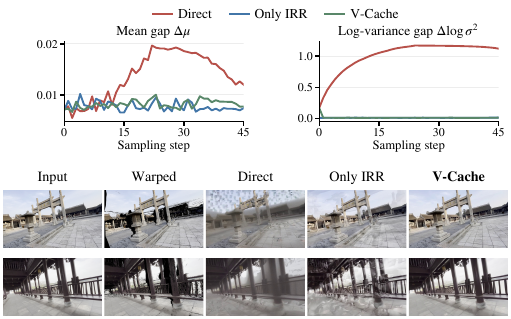}
\caption{Noise-state consistency under different reference-insertion strategies. Direct Insertion leaves gaps in the inferred-noise mean and log-variance between visible and model-generated regions, accompanied by noisy patches and broken structure. IRR reduces these gaps, and V-Cache approximates its correction while retaining continuous surfaces and clear geometric detail.}
\label{fig:noise}
\end{figure}

\paragraph{V-Cache correction}
Direct reference insertion combines an externally noised observation with a latent following the model's denoising path. To test their noise-state compatibility, we compare Direct Insertion, Only IRR, and V-Cache across denoising steps with the same inputs, target trajectories, and sampling settings (Fig.~\ref{fig:noise}). For a latent $x_i$ at noise level $\sigma_i$, the model velocity $v_i$ gives the inferred noise endpoint $\hat{\epsilon}_i=x_i+(1-\sigma_i)v_i$. We compute its mean and variance in the warped-visible and model-generated regions and plot the mean and log-variance gaps. Direct Insertion leaves persistent gaps, accompanied by noisy patches, discontinuous boundaries, and local structural errors. IRR reduces these gaps by fusing the reference in clean-estimate space and re-noising the result with one noise sample. V-Cache achieves a similar reduction without additional model evaluations (NFE), making correction more efficient.

\paragraph{Data-efficient LoRA}
Trajectory LoRA adapts only selected trajectory-sensitive blocks to strengthen the model's response to camera guidance. We use V-Cache to inject the target trajectory and compare the adapted models with UniWorld-View~\cite{uniworld}, TrajectoryCrafter~\cite{TrajCrafter}, SymphoMotion~\cite{symphomotion}, and Warp-as-History~\cite{WarpAsHistory} on trajectory accuracy, preservation of observed content, and completion of newly revealed regions. Both models use rank-128 LoRA and 50 trajectory--video training pairs. LongCat adapts its 36 most trajectory-sensitive blocks (75\%) for 1,200 updates using eight scenes; H3 adapts its 38 most sensitive blocks (approximately 75\%) for 1,500 updates using one scene. These main-comparison settings are the defaults for ablations, which vary the factor under study.

\begin{figure*}[tp]
\centering
\includegraphics[width=0.9\textwidth]{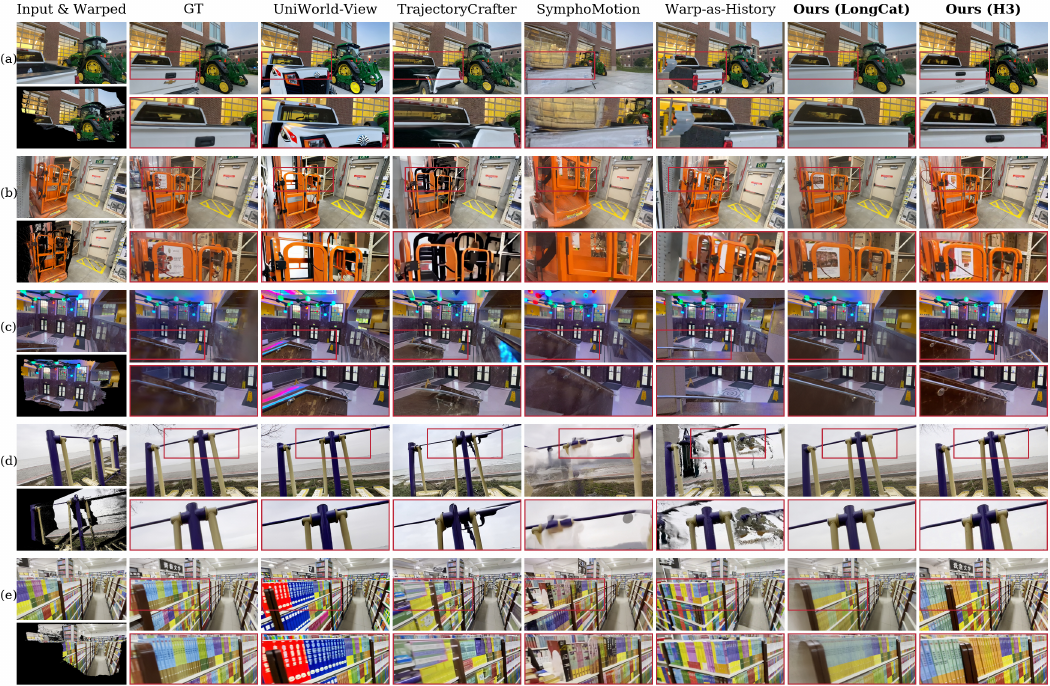}
\caption{Qualitative comparison of camera-controlled scene generation. Our LongCat and H3 models recover coherent object shapes and material detail from incomplete warped references, including thin structures and partially revealed objects. Compared with the baselines, they produce fewer rendering remnants and more continuous geometry, preserving scene appearance as the camera reveals new content.}
\label{fig:main_compare}
\end{figure*}

Both adapted models outperform the baselines on all five reconstruction metrics (Table~\ref{tab:main}). The gains in Vis-LPIPS and Disocc-LPIPS show improved preservation of observed content and completion of newly exposed regions. H3 also achieves the lowest local camera errors, combining faithful scene reconstruction with accurate camera motion. Figure~\ref{fig:main_compare} illustrates these improvements in thin structures and partially revealed objects: our models retain material detail and coherent shapes where the warped reference is incomplete, producing fewer rendering artifacts and more continuous geometry along the requested path.

\subsection{Additional Results}
\label{sec:qualitative}
\begin{figure}[!htbp]
\centering
\includegraphics[width=\columnwidth]{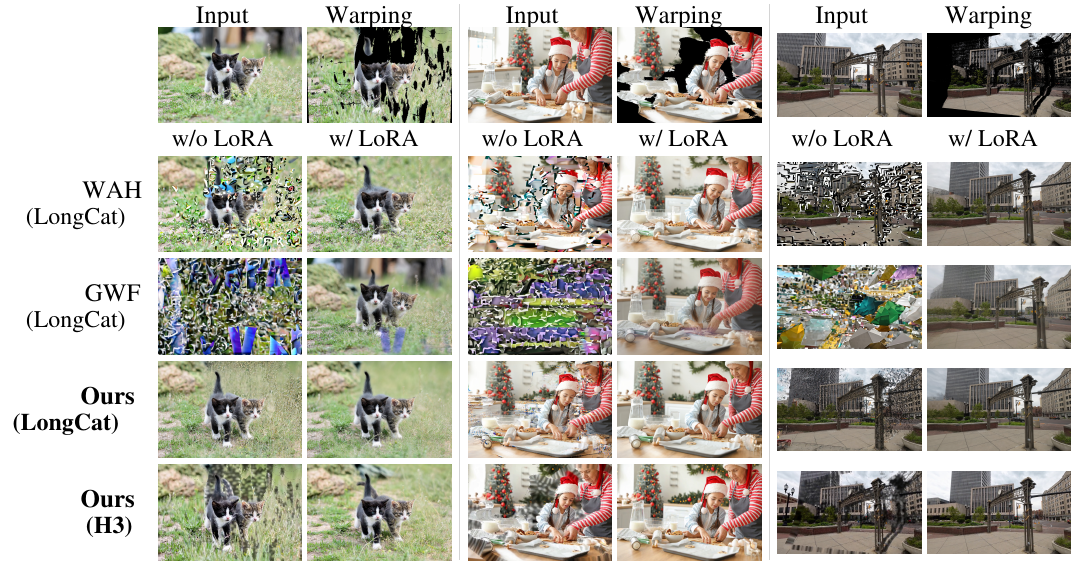}
\caption{Qualitative transfer of Trajectory LoRA across camera-control methods. Without additional fine-tuning, the same trained LongCat adapter improves surface continuity and object detail in WAH~\cite{WarpAsHistory} and GWF~\cite{GoWithTheFlow}, demonstrating transferable trajectory control.}
\label{fig:trajectory_lora_transfer}
\end{figure}

\paragraph{Generalization and transfer}
The inference-time guidance transfers directly to the U-Net-based SVD~\cite{SVD} and the Transformer-based LongCat-Video, and remains effective with different depth and optical-flow estimators (see the supplementary material). Trajectory LoRA offers a complementary form of transfer. The same LongCat adapter can be attached to Warp-as-History~\cite{WarpAsHistory} and Go-with-the-Flow~\cite{GoWithTheFlow} without further fine-tuning. It improves the model's response to the injected trajectory under both methods, reducing structural degradation and preserving more surface and object detail (Fig.~\ref{fig:trajectory_lora_transfer}). These results show that the selected blocks learn broader trajectory-control capabilities that transfer across injection interfaces.

\paragraph{Training and inference efficiency}
Training resources and inference efficiency, including accelerated generation with Trajectory LoRA, are reported in Sec.~\ref{supp:efficiency} of the supplementary material (Table~\ref{tab:supp_efficiency}).

\subsection{Ablation Studies}
\label{sec:ablation_plan}
Our ablations test how each design contributes to camera control. Following the noise-state analysis of IRR and V-Cache, we examine how FLF and DSG preserve generation quality during trajectory guidance. We then assess the effectiveness of the selected trajectory-sensitive blocks, how efficiently these blocks learn from limited scenes, and how the two training stages prepare the adapter for V-Cache-guided denoising.

\paragraph{Inference-time guidance}
IRR injects the target trajectory through the warped reference; removing this injection leaves the model without camera guidance. FLF confines reference fusion to channels aligned with the target motion, allowing the others to retain the model's appearance prediction. Removing this selection applies fusion to all channels and disrupts appearance. DSG corrects warped-reference distortions using the model's original prediction. The guided and unguided velocities differ substantially in direction, making standard CFG unstable. Table~\ref{tab:tf_ablation} shows that removing FLF or DSG reduces static and dynamic generation quality, while replacing DSG with its CFG form causes further degradation. The complete framework performs best on all four quality metrics, combining appearance-preserving fusion with correction of guidance artifacts. Qualitative component comparisons are provided in the supplementary material.

\begin{table}[!htbp]
\centering
\caption{Quantitative ablation of inference-time guidance. IRR supplies the trajectory signal; removing its reference injection leaves the model without camera guidance. With IRR fixed, we remove FLF, DSG, or both, or replace DSG with its CFG form. Both components improve generation quality, and their combination performs best on static and dynamic scenes. Bold and underlining mark the best and second-best results.}
\label{tab:tf_ablation}
\footnotesize
\setlength{\tabcolsep}{3pt}
\renewcommand{\arraystretch}{1.05}
\resizebox{0.85\columnwidth}{!}{\begin{tabular}{lrrrr}
\toprule
Method & FID$\downarrow$ & CLIP$\uparrow$ & FVD$\downarrow$ & CLIP-V$\uparrow$ \\
\midrule
w/o DSG & \underline{109.43} & 0.943 & \underline{95.69} & \underline{0.937} \\
w/o FLF & 112.69 & \underline{0.945} & 99.79 & 0.932 \\
w/o DSG and FLF & 113.12 & 0.943 & 103.17 & 0.931 \\
DSG (CFG form) & 120.91 & 0.936 & 109.10 & 0.919 \\
\midrule
Ours (Training-Free) & \textbf{96.08} & \textbf{0.948} & \textbf{93.17} & \textbf{0.938} \\
\bottomrule
\end{tabular}}
\end{table}
\begin{figure}[!tbp]
    \centering
    \includegraphics[width=0.95\columnwidth]{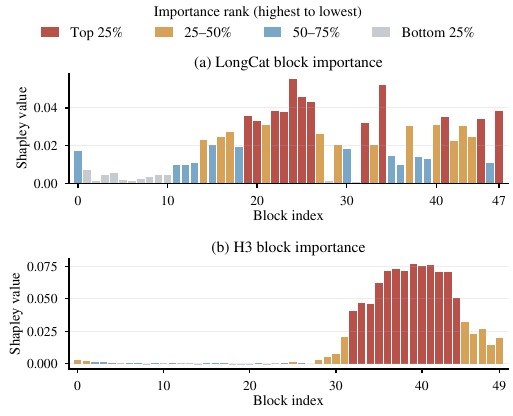}
    \caption{Shapley analysis of block sensitivity to trajectory supervision. After approximately 400 steps of probe LoRA training, contributions are estimated from 25 trajectory pairs per backbone. Sensitive blocks are dispersed through the middle and later layers of LongCat and show a more regular concentration in the later layers of H3. This variation identifies the blocks to prioritize for adaptation.}
    \label{fig:blocks}
    \end{figure}

    \begin{table*}[tp]
    \centering
    \caption{Quantitative comparison of LoRA adaptation with different block selections. Blocks are ranked by Shapley sensitivity; parentheses give the number adapted. Random uses half of the blocks, with metrics averaged over five training runs; Low response uses the least sensitive half. Other training settings follow the default configuration. Selecting roughly the most sensitive three quarters matches full-block performance and improves several reconstruction and camera metrics. Bold and underlining mark the best and second-best results within each backbone.}
    \label{tab:block_comparison}
    \footnotesize
    \setlength{\tabcolsep}{3pt}
    \renewcommand{\arraystretch}{1.05}
    \begin{minipage}[t]{0.485\textwidth}
    \centering
    \textit{LongCat}\par\smallskip
    \resizebox{\linewidth}{!}{\begin{tabular}{lrrrrrr}
    \toprule
    Selection & PSNR$\uparrow$ & SSIM$\uparrow$ & LPIPS$\downarrow$ & ATE$\downarrow$ & RPE-T$\downarrow$ & RPE-R$\downarrow$ \\
    \midrule
    All blocks (48) & \underline{18.98} & \underline{0.6140} &0.2847 & \textbf{0.01045} & \underline{0.2904} & \underline{0.1693} \\
    Selected (36) & \textbf{19.03} & \textbf{0.6176} & \textbf{0.2822} & \underline{0.01089} & \textbf{0.2806} & \textbf{0.1678} \\
    Selected (24) & 18.91 & 0.6106 & 0.2866 & 0.01163 & 0.3311 & 0.1719 \\
    Selected (12) & 18.96 & 0.6138 & \underline{0.2845} & 0.01188 & 0.3127 & 0.1696 \\
    Random (24) & 18.29 & 0.6041 & 0.2949 & 0.01241 & 0.3310 & 0.1753 \\
    Low response (24) & 17.12 & 0.5778 & 0.3336 & 0.01832 & 0.4107 & 0.1924 \\
    \bottomrule
    \end{tabular}}
    \end{minipage}\hfill
    \begin{minipage}[t]{0.485\textwidth}
    \centering
    \textit{H3}\par\smallskip
    \resizebox{\linewidth}{!}{\begin{tabular}{lrrrrrr}
    \toprule
    Selection & PSNR$\uparrow$ & SSIM$\uparrow$ & LPIPS$\downarrow$ & ATE$\downarrow$ & RPE-T$\downarrow$ & RPE-R$\downarrow$ \\
    \midrule
    All blocks (50) & \textbf{20.20} & \textbf{0.6753} & \underline{0.2783} & \underline{0.00740} & \underline{0.1742} & \underline{0.1015} \\
    Selected (38) & \underline{20.15} & 0.6741 & \textbf{0.2775} & \textbf{0.00726} & \textbf{0.1732} & \textbf{0.1003} \\
    Selected (25) & 20.01 & 0.6725 & 0.2800 & 0.00757 & 0.1821 & 0.1101 \\
    Selected (13) & 19.97 & \underline{0.6747} & 0.2810 & 0.00751 & 0.1887 & 0.1151 \\
    Random (25) & 19.44 & 0.6483 & 0.2967 & 0.00875 & 0.2256 & 0.1367 \\
    Low response (25) & 17.27 & 0.5715 & 0.3603 & 0.02339 & 0.4081 & 0.2529 \\
    \bottomrule
    \end{tabular}}
    \end{minipage}\end{table*}
\begin{figure}[!htbp]
\centering
\includegraphics[width=\columnwidth]{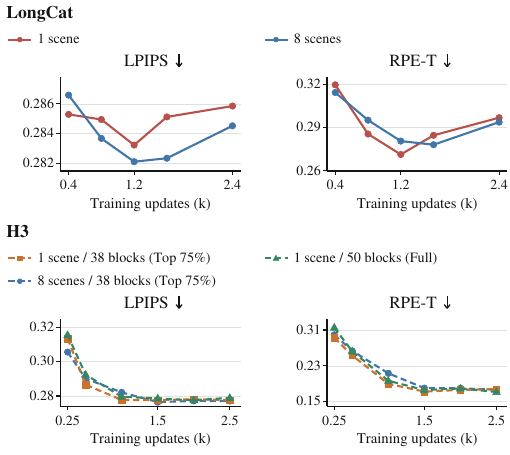}
\caption{Effect of training updates on generation quality and camera control. LongCat trains more stably with eight scenes and reaches its best perceptual quality near 1,200 updates; longer training shows signs of overfitting. H3 learns reliably from one scene. Removing LoRA from its less sensitive blocks accelerates convergence and improves perceptual quality compared with full-block adaptation. Additional updates yield diminishing gains.}
\label{fig:training}
\end{figure}

\begin{table}[!b]
\centering
\caption{Effect of training-scene diversity on camera-controlled generation. The total number of training pairs is fixed. Both backbones already produce good results after single-scene training; covering two to four scenes captures most gains from added diversity. Further expansion brings limited benefit, showing that the adapter can learn reliable camera control from few training scenes. Bold and underlining mark the best and second-best results within each backbone.}
\label{tab:scene_comparison}
\footnotesize
\setlength{\tabcolsep}{3pt}
\renewcommand{\arraystretch}{1.05}
\resizebox{0.85\columnwidth}{!}{\begin{tabular}{lrrrrrr}
\toprule
Scenes & PSNR$\uparrow$ & SSIM$\uparrow$ & LPIPS$\downarrow$ & ATE$\downarrow$ & RPE-T$\downarrow$ & RPE-R$\downarrow$ \\
\midrule
\multicolumn{7}{l}{\textit{LongCat}} \\
1 & 18.96 & 0.6108 & 0.2838 & \underline{0.01087} & \textbf{0.2723} & 0.1886 \\
2 & 19.00 & 0.6152 & 0.2851 & 0.01130 & 0.2834 & 0.1766 \\
4 & \underline{19.02} & \underline{0.6167} & \underline{0.2828} & \textbf{0.01048} & 0.2861 & \textbf{0.1638} \\
8 & \textbf{19.03} & \textbf{0.6176} & \textbf{0.2822} & 0.01089 & \underline{0.2806} & \underline{0.1678} \\
\midrule
\multicolumn{7}{l}{\textit{H3}} \\
1 & \textbf{20.15} & \textbf{0.6741} & \underline{0.2775} & 0.00726 & \underline{0.1732} & \textbf{0.1003} \\
2 & \underline{20.06} & \underline{0.6740} & 0.2817 & \textbf{0.00703} & \textbf{0.1730} & 0.1017 \\
4 & 19.86 & 0.6625 & 0.2820 & 0.00816 & 0.1833 & 0.1018 \\
8 & 19.91 & 0.6662 & \textbf{0.2761} & \underline{0.00720} & 0.1801 & \underline{0.1008} \\
\bottomrule
\end{tabular}}
\end{table}

\begin{table*}[tp]
\centering
\caption{Quantitative ablation of the two-stage training strategy and Trajectory LoRA. Standard FM trains on independently noised targets; Only warm-up and w/o warm-up isolate adaptation to recorded states and training across additional noise levels. Full combines both stages and improves reconstruction over Standard FM on both backbones, with the best camera accuracy on H3. Removing LoRA gives the poorest reconstruction.}
\label{tab:strategy}
\footnotesize
\setlength{\tabcolsep}{3pt}
\renewcommand{\arraystretch}{1.05}
\begin{minipage}[t]{0.485\textwidth}
\centering
\textit{LongCat}\par\smallskip
\resizebox{\linewidth}{!}{\begin{tabular}{lrrrrrr}
\toprule
Method & PSNR$\uparrow$ & SSIM$\uparrow$ & LPIPS$\downarrow$ & ATE$\downarrow$ & RPE-T$\downarrow$ & RPE-R$\downarrow$ \\
\midrule
Standard FM & 18.55 & 0.5477 & 0.3407 & \textbf{0.01012} & \textbf{0.2590} & \underline{0.1692} \\
Only warm-up & \underline{18.92} & 0.6062 & \underline{0.2834} & 0.01476 & 0.2985 & 0.1879 \\
w/o warm-up & 18.90 & \underline{0.6084} & 0.2857 & 0.01276 & 0.2931 & 0.1801 \\
w/o LoRA & 18.07 & 0.5271 & 0.3640 & 0.01683 & 0.2986 & 0.1964 \\
Full & \textbf{19.03} & \textbf{0.6176} & \textbf{0.2822} & \underline{0.01089} & \underline{0.2806} & \textbf{0.1678} \\
\bottomrule
\end{tabular}}
\end{minipage}\hfill
\begin{minipage}[t]{0.485\textwidth}
\centering
\textit{H3}\par\smallskip
\resizebox{\linewidth}{!}{\begin{tabular}{lrrrrrr}
\toprule
Method & PSNR$\uparrow$ & SSIM$\uparrow$ & LPIPS$\downarrow$ & ATE$\downarrow$ & RPE-T$\downarrow$ & RPE-R$\downarrow$ \\
\midrule
Standard FM & 19.55 & 0.6219 & 0.2928 & \underline{0.00767} & \underline{0.1878} & 0.1215 \\
Only warm-up & 19.87 & 0.6416 & 0.2861 & 0.00898 & 0.1904 & 0.1182 \\
w/o warm-up & \textbf{20.35} & \underline{0.6558} & \textbf{0.2771} & 0.00806 & 0.1894 & \underline{0.1106} \\
w/o LoRA & 14.85 & 0.4840 & 0.4220 & 0.01064 & 0.2458 & 0.1494 \\
Full & \underline{20.15} & \textbf{0.6741} & \underline{0.2775} & \textbf{0.00726} & \textbf{0.1732} & \textbf{0.1003} \\
\bottomrule
\end{tabular}}
\end{minipage}\end{table*}

\begin{figure*}[tp]
\centering
\includegraphics[width=0.85\textwidth]{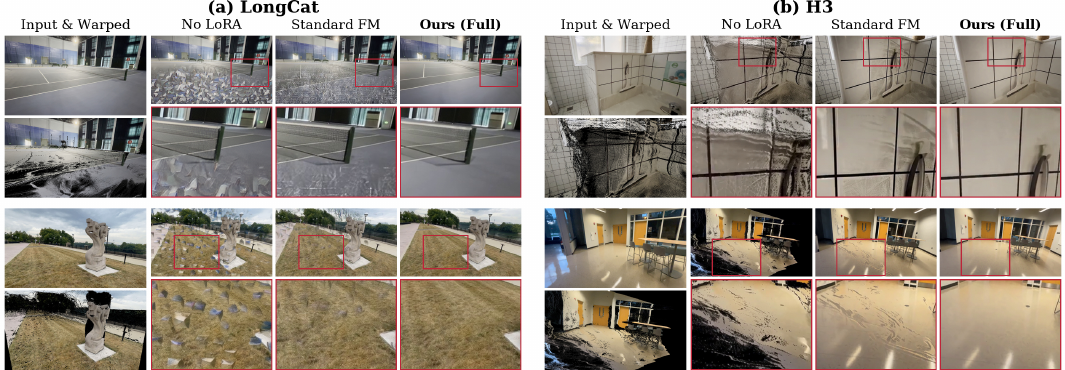}
\caption{Qualitative comparison of Trajectory LoRA training strategies. For LongCat (a) and H3 (b), two examples per backbone place the input and warped reference beside No LoRA, Standard FM, and the full method. The full strategy reduces fragmented geometry and residual rendering artifacts, producing more continuous surfaces and finer scene detail than No LoRA and Standard FM. Red boxes and enlarged crops compare corresponding regions.}
\label{fig:strategy}
\end{figure*}

\paragraph{Trajectory-sensitive block selection}
We use Shapley analysis~\cite{Shapley1953} to measure how strongly each self-attention block responds to trajectory supervision. A block's score averages its marginal contribution to the full probe LoRA's velocity response across different combinations of active adapters. After brief probe LoRA training on each backbone, Fig.~\ref{fig:blocks} reveals clear differences in sensitivity. LongCat's responsive blocks are dispersed through the middle and later layers, while H3 shows a more regular concentration in later layers.

To test whether this sensitivity guides effective adaptation, we train different proportions of the highest-scoring blocks and compare them with full-block adaptation. Controls use the least responsive half or a random half, with random-selection results averaged over five independent training runs. All LoRA ablations use the same 100 test cases: five random camera trajectories from each of 20 randomly selected DL3DV scenes, disjoint from all training scenes. Other training settings follow the default configuration.

Adapting roughly the most sensitive three quarters of the blocks matches or improves full-block performance (Table~\ref{tab:block_comparison}). LongCat's 36 selected blocks improve reconstruction and local camera accuracy, while H3's 38 selected blocks retain comparable reconstruction and reduce camera errors. These gains support leaving low-sensitivity blocks unadapted to preserve pretrained texture and appearance priors. The selected H3 blocks also reach good performance sooner than full-block adaptation (Fig.~\ref{fig:training}). At the same block count, selecting the sensitive half improves reconstruction over random selection on both backbones, while low-response selection performs worst. These comparisons validate our block-selection strategy.

\paragraph{Training-scene diversity}
With adaptation focused on trajectory-sensitive blocks, we examine how much scene diversity is needed for reliable camera control. We draw a fixed total of 50 trajectory--video pairs from one, two, four, or eight DL3DV scenes, keeping other training settings unchanged. Both backbones already achieve good reconstruction and control with one scene (Table~\ref{tab:scene_comparison}). Added diversity improves LongCat's reconstruction and rotation accuracy, while H3 gains mainly in translation accuracy. Most gains are reached with two to four scenes; extending coverage to eight brings limited further benefit. These results show that the training strategy uses scene supervision efficiently, enabling reliable trajectory control and high-quality generation on unseen scenes from few examples.

\paragraph{Training updates}
We next vary training updates while fixing each backbone's training data and other settings. Figure~\ref{fig:training} tracks the eight-scene LongCat and single-scene H3 configurations, with additional scene-count comparisons and full-block adaptation on H3. LongCat trains more stably with eight scenes and reaches its best perceptual quality near 1,200 updates; longer training brings little benefit and eventually shows signs of overfitting. H3 learns reliably from one scene. Removing LoRA from its less sensitive blocks accelerates convergence and improves perceptual quality over full-block adaptation. Later updates refine camera accuracy as perceptual quality stabilizes, with diminishing gains. Thus, the sensitive blocks learn to use trajectory guidance within relatively few updates.

\paragraph{Training strategy}
We compare Standard FM, Only warm-up, and w/o warm-up to isolate the effects of the training strategies, using 1,200 updates for LongCat and 1,500 for H3. The adapter must learn to denoise states modified by reference insertion and cached correction. Anchor warm-up first adapts it to intermediate states recorded during guided sampling; the second stage uses the warmed-up model and these states to construct guided inputs at additional noise levels. Table~\ref{tab:strategy} and Fig.~\ref{fig:strategy} compare these variants with Full and w/o LoRA under fixed inference settings for each backbone. Standard FM uses independently noised targets. Only warm-up trains on recorded states without extending the noise range, while w/o warm-up skips initial adaptation but still constructs guided inputs from intermediate states at additional noise levels. The full strategy gives the best overall results across quantitative and qualitative comparisons, improving reconstruction over Standard FM on both backbones and achieving the best camera accuracy on H3. Removing LoRA gives the poorest reconstruction. Although Standard FM has low errors on some camera metrics, its training inputs differ substantially from inference-time guided states. This gap prevents reliable use of trajectory guidance, leaving holes, cracks, and rendering artifacts. Training on states that reproduce guided denoising restores continuous surfaces and finer detail.

\section{Conclusion}
\label{sec:conclusion}

We presented WorldForge V2 for accurate camera control while preserving the generative capabilities of pretrained video models. Our training-free guidance uses warped references to iteratively refine generation along the requested camera path, with motion-aware channel selection and self-correction preserving scene appearance and object motion. Building on this guidance, \vcache{} reuses denoising predictions to approximate the refinement correction, reducing sampling cost. Trajectory LoRA extends motion localization from latent channels to network blocks and combines selective adaptation with two-stage training, enabling pretrained video models to learn reliable trajectory control from very few trajectory--video pairs. Quantitative and qualitative evaluations across multiple datasets demonstrate state-of-the-art generation quality and trajectory accuracy in camera-controlled video generation. Future work will explore long-horizon, real-time streaming generation with camera control to support the development of interactive world models.

\FloatBarrier

\FloatBarrier
\clearpage

\begingroup
\setcounter{section}{0}
\setcounter{subsection}{0}
\setcounter{figure}{0}
\setcounter{table}{0}
\setcounter{equation}{0}
\setcounter{secnumdepth}{2}
\setcounter{tocdepth}{2}
\renewcommand{\thesection}{\arabic{section}}
\renewcommand{\thesubsection}{\thesection.\arabic{subsection}}
\renewcommand{\thefigure}{S\arabic{figure}}
\renewcommand{\thetable}{S\arabic{table}}
\renewcommand{\theequation}{S\arabic{equation}}
\renewcommand{\theHsection}{supp.\arabic{section}}
\renewcommand{\theHsubsection}{\theHsection.\arabic{subsection}}
\renewcommand{\theHfigure}{supp.\arabic{figure}}
\renewcommand{\theHtable}{supp.\arabic{table}}
\renewcommand{\theHequation}{supp.\arabic{equation}}
\setlength{\parskip}{0pt}
\setlength{\parindent}{1em}
\setlength{\abovedisplayskip}{4pt plus 2pt minus 1pt}
\setlength{\belowdisplayskip}{4pt plus 2pt minus 1pt}
\setlength{\abovedisplayshortskip}{0pt plus 2pt}
\setlength{\belowdisplayshortskip}{4pt plus 2pt minus 1pt}
\setlength{\textfloatsep}{8pt plus 2pt minus 2pt}
\setlength{\dbltextfloatsep}{8pt plus 2pt minus 2pt}
\setlength{\floatsep}{8pt plus 2pt minus 2pt}
\setlength{\dblfloatsep}{8pt plus 2pt minus 2pt}
\setlength{\intextsep}{6pt plus 2pt minus 2pt}
\setlength{\abovecaptionskip}{3pt}
\setlength{\belowcaptionskip}{2pt}
\renewcommand{\topfraction}{0.95}
\renewcommand{\dbltopfraction}{0.95}
\renewcommand{\bottomfraction}{0.85}
\renewcommand{\textfraction}{0.05}
\renewcommand{\floatpagefraction}{0.75}
\renewcommand{\dblfloatpagefraction}{0.75}
\setcounter{topnumber}{5}
\setcounter{bottomnumber}{4}
\setcounter{dbltopnumber}{4}
\setcounter{totalnumber}{10}
\clubpenalty=10000
\widowpenalty=10000
\raggedbottom

\makeatletter
\renewcommand{\thesectiondis}{\thesection.}
\renewcommand{\thesubsectiondis}{\thesubsection.}
\def\@IEEEsectpunct{}
\def\@seccntformat#1{\csname the#1dis\endcsname\hskip 0.6em}
\renewcommand{\section}{\@startsection{section}{1}{\z@}{-2.2ex plus -0.4ex minus -0.2ex}{0.8ex}{\normalfont\fontsize{11}{12.5}\selectfont\bfseries}}
\renewcommand{\subsection}{\@startsection{subsection}{2}{\z@}{-1.8ex plus -0.3ex minus -0.2ex}{0.6ex}{\normalfont\normalsize\bfseries}}
\setlength{\@fptop}{0pt}
\setlength{\@fpsep}{8pt}
\setlength{\@fpbot}{0pt plus 1fil}
\setlength{\@dblfptop}{0pt}
\setlength{\@dblfpsep}{8pt}
\setlength{\@dblfpbot}{0pt plus 1fil}
\renewcommand{\fnum@figure}{Figure~\thefigure}
\renewcommand{\fnum@table}{Table~\thetable}
\long\def\@makecaption#1#2{\vskip\abovecaptionskip
  {\normalfont\fontsize{8.5}{10}\selectfont
   \noindent #1. #2\par}\vskip\belowcaptionskip}
\def\Hy@bookmarkstype{stoc}
\let\supp@addcontentsline\addcontentsline
\def\supp@tocname{toc}
\renewcommand{\addcontentsline}[3]{\def\supp@extension{#1}\ifx\supp@extension\supp@tocname
    \supp@addcontentsline{stoc}{#2}{#3}\else
    \supp@addcontentsline{#1}{#2}{#3}\fi}
\renewcommand{\l@section}[2]{\addvspace{2pt}{\bfseries\@dottedtocline{1}{0em}{1.3em}{#1}{#2}}}
\renewcommand{\l@subsection}{\@dottedtocline{2}{1.3em}{2.1em}}
\newcommand{\suppcontents}{\section*{Contents}{\small\@starttoc{stoc}}\vspace{0.6em}}
\makeatother

\twocolumn[{\begin{center}
{\Large\bfseries WorldForge V2: Eliciting Camera-Trajectory Alignment\\
in Video Diffusion Models\par}
\vspace{6pt}
{\large\bfseries Supplementary Material\par}
\end{center}
\vspace{8pt}
}]
\suppcontents

\section{Equivalence between Diffusion and Flow Models}
\label{supp:training-free}

Our camera guidance uses the model's estimate of the clean video latent. This estimate can be obtained from either a diffusion noise prediction or a Flow Matching velocity prediction. We derive their relationship by expressing the Flow Matching path~\citep{Flow} through the noise schedule of a diffusion model~\citep{PFODE}.

Let $\mathbf{x}_0$ be drawn from the data distribution $q_0$. A diffusion model~\citep{DDPM,PFODE} gradually adds Gaussian noise to the data, giving the conditional distribution
\begin{equation}
q_t(\mathbf{x}_t\mid\mathbf{x}_0)
=\mathcal{N}(\mathbf{x}_t\mid\alpha_t\mathbf{x}_0,\sigma_t^2\mathbf{I}).
\label{eq:supp_distribution}
\end{equation}

Equivalently, a sample at time $t$ can be written as~\citep{PFODE}
\begin{equation}
\mathbf{x}_t=\alpha_t\mathbf{x}_0+\sigma_t\bm{\epsilon},
\qquad \bm{\epsilon}\sim\mathcal{N}(\mathbf{0},\mathbf{I}).
\label{eq:supp_xt}
\end{equation}

The noise schedules $\alpha_t$ and $\sigma_t$ control the signal and noise levels: $\alpha_t$ typically decreases as $\sigma_t$ increases. Variance-preserving diffusion satisfies $\alpha_t^2+\sigma_t^2=1$~\citep{DDPM,PFODE}. More generally, the Gaussian transitions in Eq.~\eqref{eq:supp_distribution} correspond to the forward stochastic differential equation (SDE)
\begin{equation}
\mathrm{d}\mathbf{x}_t=f(t)\mathbf{x}_t\mathrm{d}t+g(t)\mathrm{d}\mathbf{w}_t,
\qquad \mathbf{x}_0\sim q_0,
\label{eq:supp_sde}
\end{equation}
where $\mathbf{w}_t$ is a standard Wiener process, and the drift and diffusion coefficients are determined by the noise schedules~\citep{PFODE}:
\begin{equation}
\begin{aligned}
f(t)&=\frac{\mathrm{d}\log\alpha_t}{\mathrm{d}t},\\
g^2(t)&=\frac{\mathrm{d}\sigma_t^2}{\mathrm{d}t}
-2\frac{\mathrm{d}\log\alpha_t}{\mathrm{d}t}\sigma_t^2.
\end{aligned}
\label{eq:supp_ft}
\end{equation}

Generation reverses this process from noise to data. The associated probability-flow ordinary differential equation (ODE) follows the same marginal distributions as the SDE~\citep{PFODE}. Denoting the marginal density of $\mathbf{x}_t$ by $p_t$, we have
\begin{equation}
\frac{\mathrm{d}\mathbf{x}_t}{\mathrm{d}t}
=f(t)\mathbf{x}_t-\frac{1}{2}g^2(t)\nabla_{\mathbf{x}_t}\log p_t(\mathbf{x}_t).
\label{eq:supp_ode}
\end{equation}

The noise predictor $\bm{\epsilon}_\theta$ approximates the score as $\nabla_{\mathbf{x}_t}\log p_t(\mathbf{x}_t)\approx-\bm{\epsilon}_\theta(\mathbf{x}_t,t)/\sigma_t$. Substituting it into Eq.~\eqref{eq:supp_ode} gives the sampling ODE~\citep{PFODE}:
\begin{equation}
\frac{\mathrm{d}\mathbf{x}_t}{\mathrm{d}t}
=f(t)\mathbf{x}_t+\frac{g^2(t)}{2\sigma_t}\bm{\epsilon}_\theta(\mathbf{x}_t,t).
\label{eq:supp_ode_simplified}
\end{equation}

For Flow Matching~\citep{Flow}, we consider the path obtained by linearly interpolating between a data sample and Gaussian noise:
\begin{equation}
\mathbf{x}_t=(1-t)\mathbf{x}_0+t\bm{\epsilon},
\qquad \bm{\epsilon}\sim\mathcal{N}(\mathbf{0},\mathbf{I}),
\label{eq:supp_fm}
\end{equation}
where $t\in[0,1]$. Comparing Eq.~\eqref{eq:supp_fm} with Eq.~\eqref{eq:supp_xt} gives $\alpha_t=1-t$ and $\sigma_t=t$. Substitution into Eq.~\eqref{eq:supp_ft} yields
\begin{equation}
f_{\mathrm{FM}}(t)=\frac{\mathrm{d}\log(1-t)}{\mathrm{d}t}=-\frac{1}{1-t},
\label{eq:supp_fm_drift}
\end{equation}
\begin{equation}
g^2_{\mathrm{FM}}(t)=\frac{\mathrm{d}(t^2)}{\mathrm{d}t}+\frac{2t^2}{1-t}
=\frac{2t}{1-t}.
\label{eq:supp_fm_diffusion}
\end{equation}

With an ideal noise predictor, $\bm{\epsilon}_\theta(\mathbf{x}_t,t)=\mathbb{E}[\bm{\epsilon}\mid\mathbf{x}_t]$. Using the coefficients above in Eq.~\eqref{eq:supp_ode_simplified}, we obtain
\begin{equation}
\begin{aligned}
\frac{\mathrm{d}\mathbf{x}_t}{\mathrm{d}t}
&=\frac{\bm{\epsilon}_\theta(\mathbf{x}_t,t)-\mathbf{x}_t}{1-t}\\
&=\mathbb{E}[\bm{\epsilon}-\mathbf{x}_0\mid\mathbf{x}_t],
\qquad 0<t<1.
\end{aligned}
\label{eq:supp_diffusionflow}
\end{equation}

The conditional path in Eq.~\eqref{eq:supp_fm} has derivative $\bm{\epsilon}-\mathbf{x}_0$, whose conditional expectation gives the marginal Flow Matching velocity. Thus, the Flow Matching path considered here corresponds to the diffusion schedule $\alpha_t=1-t$, $\sigma_t=t$~\citep{Flow}. For corresponding noise and velocity predictions, the clean latent used by our guidance is
\begin{equation}
\begin{aligned}
\hat{\mathbf{x}}_0
&=\frac{\mathbf{x}_t-t\bm{\epsilon}_\theta(\mathbf{x}_t,t)}{1-t}\\
&=\mathbf{x}_t-t\mathbf{v}_\theta(\mathbf{x}_t,t),
\qquad 0<t<1.
\end{aligned}
\label{eq:supp_clean_estimate}
\end{equation}

\section{Evaluation Protocol}
\label{supp:evaluation}

Camera control requires the generated video to follow the target path while preserving the scene. We evaluate these two aspects through visual quality and trajectory accuracy on the training-free benchmarks, and use the rendered DL3DV targets to further examine reconstruction of observed content and completion of newly exposed regions.

\subsection{Data and Comparison Protocol}
\label{supp:data_protocol}

The training-free static benchmark contains more than 70 source views from over 40 scenes, including LLFF, Tanks and Temples, Mip-NeRF 360, and real or generated images. The dynamic benchmark contains more than 50 trajectory-controlled outputs from over 30 source videos, including DAVIS, cinematic sequences, and videos generated by video diffusion models.

The learned-control experiments use DL3DV~\citep{DL3DV}, with disjoint training and test scenes. The main comparison contains 200 cases, with five random camera paths for each of 40 scenes. All LoRA ablations use the same 100 cases, comprising five random paths from each of 20 randomly selected scenes. Sequences contain 25--93 frames, with lengths adapted to each method's supported input format. Across the benchmarks, methods within each comparison share the inputs, target trajectories, and evaluation cases.

\subsection{Static Scene Evaluation}
\label{supp:static_metrics}

Fr\'echet Inception Distance (FID)~\citep{FID} measures visual quality by comparing the feature distributions of real and generated images. We resize images to $299\times299$ and extract 2048-dimensional pool3 features from an ImageNet-pretrained Inception-V3 model. The Fr\'echet distance between the Gaussian distributions fitted to these features is
\begin{equation}
\begin{aligned}
\mathrm{FID}={}&\|\mu_r-\mu_g\|_2^2\\
&+\mathrm{Tr}\!\left(\Sigma_r+\Sigma_g-2(\Sigma_r\Sigma_g)^{1/2}\right),
\end{aligned}
\label{eq:supp_fid}
\end{equation}
where $\mu_r,\mu_g$ and $\Sigma_r,\Sigma_g$ are the means and covariances of the real and generated features, computed over the evaluation set. The reference distribution aggregates ground-truth images from the static scenes. For in-the-wild scenes without ground truth, we use manually curated reference sets.

CLIP similarity~\citep{CLIP} complements FID by measuring semantic agreement between corresponding real and generated images. We use CLIP ViT-B/32, pretrained on 400 million image--text pairs, to extract L2-normalized 512-dimensional features. Their average cosine similarity over $N$ image pairs is
\begin{equation}
\mathrm{CLIP}_{\mathrm{sim}}=\frac{1}{N}\sum_{i=1}^{N}\cos(f_{r,i},f_{g,i}),
\label{eq:supp_clip}
\end{equation}
where $f_{r,i}$ and $f_{g,i}$ are the features of the $i$-th real and generated image pair.

\subsection{Dynamic Scene Evaluation}
\label{supp:dynamic_metrics}

For dynamic scenes, Fr\'echet Video Distance (FVD)~\citep{FVD} compares real and generated videos using spatiotemporal features. We resize I3D inputs to $224\times224$ and extract a 1024-dimensional feature for each clip from the global average pooling layer of a Kinetics-pretrained I3D network. As with FID, we fit Gaussian distributions to the real and generated features and compute
\begin{equation}
\begin{aligned}
\mathrm{FVD}={}&\|\mu_r-\mu_g\|_2^2\\
&+\mathrm{Tr}\!\left(\Sigma_r+\Sigma_g-2(\Sigma_r\Sigma_g)^{1/2}\right),
\end{aligned}
\label{eq:supp_fvd}
\end{equation}
where the means and covariances are estimated from clip-level I3D features over the evaluation set. Reference clips are extracted from DAVIS and cinematic sequences.

Video CLIP similarity ($\mathrm{CLIP\text{-}V}_{\mathrm{sim}}$) measures frame-level semantic agreement. We average the CLIP similarity of corresponding frames within each video pair, then average over all $M$ pairs:
\begin{equation}
\mathrm{CLIP\text{-}V}_{\mathrm{sim}}=
\frac{1}{M}\sum_{j=1}^{M}\frac{1}{T_j}\sum_{t=1}^{T_j}
\cos(f_{r,j,t},f_{g,j,t}),
\label{eq:supp_clip_video}
\end{equation}
where $T_j$ is the number of evaluated frames in pair $j$, and $f_{r,j,t},f_{g,j,t}$ are the corresponding CLIP features. Video evaluation uses 16 uniformly sampled frames, including the first and last frames.

\subsection{Reconstruction and Completion}
\label{supp:reconstruction_metrics}

The rendered DL3DV targets allow us to compare generated content with the scene along each camera path. We use PSNR for pixel agreement, SSIM for structural agreement, and LPIPS~\citep{lpips} for perceptual similarity. Frames are evaluated on a common $640\times360$ image area, excluding the input frame.

To distinguish preservation of observed content from completion of newly exposed regions, we also report LPIPS for these regions separately. Let $D_i(p)$ be the spatial LPIPS error map for frame $i$ and $M_i(p)$ its rendered visibility mask. The regional scores are
\begin{equation}
\begin{aligned}
\mathrm{Vis}_i&=\frac{\sum_p M_i(p)D_i(p)}{\sum_p M_i(p)},\\
\mathrm{Disocc}_i&=\frac{\sum_p[1-M_i(p)]D_i(p)}{\sum_p[1-M_i(p)]}.
\end{aligned}
\label{eq:supp_regional_lpips}
\end{equation}

Vis-LPIPS evaluates the observed region, and Disocc-LPIPS evaluates the newly exposed region. Empty regions are omitted. We average scores over valid frames within each case, then over cases within each scene, and finally over scenes with equal weight.

\subsection{Camera-Trajectory Accuracy}
\label{supp:camera_metrics}

For the training-free benchmarks, we use \emph{evo} to compute root-mean-square errors (RMSE). A global positive-scale similarity transform aligns the estimated trajectory to the target by correcting scale, rotation, and translation. We denote the aligned camera position and rotation at frame $i$ by $\tilde{\mathbf{t}}_{\mathrm{est},i}$ and $\tilde{\mathbf{R}}_{\mathrm{est},i}$. Absolute Trajectory Error (ATE) measures global position agreement over the $n$ camera poses:

\begin{equation}
\begin{aligned}
\mathrm{ATE}_i&=\|\mathbf{t}_{\mathrm{ref},i}-\tilde{\mathbf{t}}_{\mathrm{est},i}\|_2,\\
\mathrm{ATE}&=\sqrt{\frac{1}{n}\sum_{i=1}^{n}\mathrm{ATE}_i^2}.
\end{aligned}
\label{eq:supp_tf_ate}
\end{equation}

Local motion accuracy is measured by Relative Pose Error in translation (RPE-T). With an index gap of $\Delta=1$, the relative poses between adjacent frames are
\begin{equation}
\begin{aligned}
\Delta\mathbf{T}_{\mathrm{ref},i}
&=\mathbf{T}_{\mathrm{ref},i}^{-1}\mathbf{T}_{\mathrm{ref},i+1},\\
\Delta\mathbf{T}_{\mathrm{est},i}
&=\tilde{\mathbf{T}}_{\mathrm{est},i}^{-1}\tilde{\mathbf{T}}_{\mathrm{est},i+1}.
\end{aligned}
\label{eq:supp_relative_poses}
\end{equation}

Denoting their translation components by $\Delta\mathbf{t}_{\mathrm{ref},i}$ and $\Delta\mathbf{t}_{\mathrm{est},i}$, we compute the per-step translation error and its RMSE as
\begin{equation}
\begin{aligned}
\mathrm{RPE\text{-}T}_i
&=\|\Delta\mathbf{t}_{\mathrm{ref},i}-\Delta\mathbf{t}_{\mathrm{est},i}\|_2,\\
\mathrm{RPE\text{-}T}
&=\sqrt{\frac{1}{n-1}\sum_{i=1}^{n-1}\mathrm{RPE\text{-}T}_i^2}.
\end{aligned}
\label{eq:supp_tf_rpet}
\end{equation}

Relative Pose Error in rotation (RPE-R) measures the corresponding error in orientation changes. The relative rotations between adjacent frames are
\begin{equation}
\begin{aligned}
\Delta\mathbf{R}_{\mathrm{ref},i}
&=\mathbf{R}_{\mathrm{ref},i}^{-1}\mathbf{R}_{\mathrm{ref},i+1},\\
\Delta\mathbf{R}_{\mathrm{est},i}
&=\tilde{\mathbf{R}}_{\mathrm{est},i}^{-1}\tilde{\mathbf{R}}_{\mathrm{est},i+1}.
\end{aligned}
\label{eq:supp_relative_rotations}
\end{equation}

We compute the angular error in degrees and take its RMSE over the trajectory:
{\small
\begin{equation}
\begin{aligned}
\mathrm{RPE\text{-}R}_i
&=\arccos\!\left(\frac{\mathrm{Tr}(\Delta\mathbf{R}_{\mathrm{ref},i}^{\mathsf T}\Delta\mathbf{R}_{\mathrm{est},i})-1}{2}\right)\frac{180}{\pi},\\
\mathrm{RPE\text{-}R}
&=\sqrt{\frac{1}{n-1}\sum_{i=1}^{n-1}\mathrm{RPE\text{-}R}_i^2}.
\end{aligned}
\label{eq:supp_tf_rper}
\end{equation}
}

For DL3DV, we also align recovered cameras to the target with a positive-scale similarity transform. To compare paths of different lengths, we normalize translation errors by the reference motion. ATE uses normalized position RMSE, RPE-T divides the sum of translation errors by the sum of reference translation magnitudes, and RPE-R averages angular errors over adjacent poses.

Let $\hat{\mathbf{c}}_i$ and $\mathbf{c}_i$ be the aligned estimated and target camera centers, and $\Delta\hat{\mathbf{T}}_i$, $\Delta\mathbf{T}_i$ their adjacent relative poses. With reference path length $L=\sum_{i=1}^{n-1}\|\mathbf{c}_{i+1}-\mathbf{c}_i\|_2$, ATE is
\begin{equation}
\mathrm{ATE}=\frac{1}{L}\sqrt{\frac{1}{n}\sum_{i=1}^{n}\|\hat{\mathbf{c}}_i-\mathbf{c}_i\|_2^2}.
\label{eq:supp_dl3dv_ate}
\end{equation}

Writing $\Delta\mathbf{t}_i$ and $\Delta\mathbf{R}_i$ for the relative translation and rotation, the relative pose errors are
\begin{equation}
\mathrm{RPE\text{-}T}
=\frac{\sum_{i=1}^{n-1}\|\Delta\hat{\mathbf{t}}_i-\Delta\mathbf{t}_i\|_2}
{\sum_{i=1}^{n-1}\|\Delta\mathbf{t}_i\|_2},
\label{eq:supp_dl3dv_rpet}
\end{equation}
\begin{equation}
\mathrm{RPE\text{-}R}=\frac{1}{n-1}\sum_{i=1}^{n-1}
\angle(\Delta\mathbf{R}_i^{\mathsf T}\Delta\hat{\mathbf{R}}_i),
\label{eq:supp_dl3dv_rper}
\end{equation}
with rotation angles in degrees. The input frame is excluded from alignment and scoring. We use GLOMAP~\citep{glomap} to recover poses for the main comparison and H3 studies, and Pi3X~\citep{Pi3} for the LongCat training ablations. All methods within each comparison use the same pose estimator and aggregation.

\section{Implementation Details}
\label{supp:implementation}

\subsection{FLF Motion Estimation and Channel Scoring}
\label{supp:flf_implementation}

Flow-Gated Latent Fusion (FLF) uses motion agreement to select the latent channels that receive trajectory guidance. At each denoising step $i$, we compare the motion in each predicted channel with that in the rendered reference and assign a score $S_i^c$. Channels with high scores receive the reference fusion, while the others retain the model's prediction to preserve appearance details.

For each channel $c$, we apply Farneb\"ack optical flow~\citep{OpticaFlow} between adjacent frames of the predicted clean latent $\hat{\mathbf{x}}_0^{(i)}$ and the trajectory latent $\mathbf{x}_{\mathrm{traj}}$. This gives predicted flow $\mathcal{F}_{\mathrm{pred}}^{(i,c)}$ and reference flow $\mathcal{F}_{\mathrm{ref}}^{(i,c)}$, with each flow vector describing horizontal and vertical displacement. For a latent video with $T_l$ frames, the frame-pair index is $\tau=1,\ldots,T_l-1$. We compare the flows over valid locations in the rendered visibility mask, $\Omega^{(i,c)}=\{(x,y,\tau)\mid\mathbf{M}^{(c)}(x,y,\tau)=1\}$, where $(x,y)$ are spatial coordinates.

We measure motion agreement through displacement error, angular error, and the proportion of large mismatches. The masked endpoint error (M-EPE) averages the distance between the two flow vectors over valid locations:
{\small
\begin{equation}
e(x,y,\tau)=\left\|\mathcal{F}_{\mathrm{pred}}^{(i,c)}(x,y,\tau)
-\mathcal{F}_{\mathrm{ref}}^{(i,c)}(x,y,\tau)\right\|_2.
\label{eq:supp_endpoint_error}
\end{equation}
}
\begin{equation}
\mathrm{M\text{-}EPE}^{(i,c)}
=\frac{1}{|\Omega^{(i,c)}|}\sum_{(x,y,\tau)\in\Omega^{(i,c)}}e(x,y,\tau).
\label{eq:supp_mepe}
\end{equation}

To compare flow directions, we compute their cosine similarity at each valid location and average the corresponding angles in degrees to obtain the masked angular error (M-AE):
{\small
\begin{equation}
s(x,y,\tau)=\frac{\mathcal{F}_{\mathrm{pred}}^{(i,c)}(x,y,\tau)\cdot\mathcal{F}_{\mathrm{ref}}^{(i,c)}(x,y,\tau)}
{\|\mathcal{F}_{\mathrm{pred}}^{(i,c)}(x,y,\tau)\|_2\,\|\mathcal{F}_{\mathrm{ref}}^{(i,c)}(x,y,\tau)\|_2}.
\label{eq:supp_flow_cosine}
\end{equation}
}
\begin{equation}
\mathrm{M\text{-}AE}^{(i,c)}
=\frac{180}{\pi|\Omega^{(i,c)}|}\sum_{(x,y,\tau)\in\Omega^{(i,c)}}\arccos s(x,y,\tau).
\label{eq:supp_mae}
\end{equation}

The outlier percentage $\mathrm{F}^{(i,c)}$ measures the proportion of valid locations whose endpoint error exceeds 3 pixels or whose relative error $e(x,y,\tau)/\|\mathcal{F}_{\mathrm{ref}}^{(i,c)}(x,y,\tau)\|_2$ exceeds 5\%. It captures large local mismatches alongside the average displacement and angular errors.

We normalize these three errors to $[0,1]$ before combining them:
\begin{equation}
\begin{aligned}
E_i^c&=\min(\mathrm{M\text{-}EPE}^{(i,c)}/n_{\mathrm{E}},1),\\
A_i^c&=\min(\mathrm{M\text{-}AE}^{(i,c)}/n_{\mathrm{A}},1),\\
O_i^c&=\min(\mathrm{F}^{(i,c)}/n_{\mathrm{F}},1).
\end{aligned}
\label{eq:supp_flf_normalization}
\end{equation}

Subtracting each normalized error from one gives a score that increases with motion agreement. The weighted motion score is
\begin{equation}
S_i^c=\gamma_{\mathrm{E}}(1-E_i^c)+\gamma_{\mathrm{A}}(1-A_i^c)
+\gamma_{\mathrm{O}}(1-O_i^c).
\label{eq:supp_flf_score}
\end{equation}

We set $(n_{\mathrm{E}},n_{\mathrm{A}},n_{\mathrm{F}})=(10,30^\circ,0.5)$ and $(\gamma_{\mathrm{E}},\gamma_{\mathrm{A}},\gamma_{\mathrm{O}})=(0.4,0.4,0.2)$, with $\sum_k\gamma_k=1$. At each step, the selection threshold follows the distribution of channel scores:
\begin{equation}
\mathcal{C}_i=\{c\mid S_i^c\ge\delta_i\},
\qquad \delta_i=\mu_i^S-\lambda_i\sigma_i^S,
\label{eq:supp_flf_selection}
\end{equation}
where $\mu_i^S$ and $\sigma_i^S$ are the mean and standard deviation of the channel scores, and $\lambda_i$ controls the threshold. Channels in $\mathcal{C}_i$ receive the reference fusion.

\subsection{Hyperparameter Settings}
\label{supp:hyperparams}

For training-free control with Wan2.1 I2V-14B, we use a 50-step UniPC sampler and apply Intra-Step Recursive Refinement (IRR) during the first 20 steps by default. The FLF and Dual-Path Self-Corrective Guidance (DSG) coefficients are given in Table~\ref{tab:supp_hyperparams}. Farneb\"ack flow estimation uses \texttt{pyr\_scale}=0.5, \texttt{levels}=3, and \texttt{winsize}=15~\citep{OpticaFlow}.

\begin{table}[!t]
\centering
\caption{Default guidance coefficients for the Wan2.1 training-free experiments. The flow weights combine displacement, direction, and outlier errors; $\lambda_i$ controls the adaptive channel threshold in Eq.~\eqref{eq:supp_flf_selection}, and $\rho$ sets the strength of DSG correction.}
\label{tab:supp_hyperparams}
\footnotesize\setlength{\tabcolsep}{4pt}
\renewcommand{\arraystretch}{1.1}
\begin{tabular}{@{}lcl@{}}
\toprule
Coefficient & Value & Description\\
\midrule
$\gamma_k$ & $(0.4,0.4,0.2)$ & Optical-flow metric weights\\
$\lambda_i$ & $0.65$ & FLF threshold coefficient\\
$\rho$ & $4.0$ & DSG guidance scale\\
\bottomrule
\end{tabular}
\end{table}

Early latent states are too noisy for reliable flow estimation, so we disable channel filtering and inject the reference into all channels during steps 0--5. As the motion estimates become more reliable, we allow at most two of Wan2.1's 16 channels to retain the original prediction during steps 6--10; at least 14 channels therefore continue to receive the reference. During steps 11--19, we relax this limit to six channels, giving the model more freedom to recover texture while maintaining trajectory guidance in the remaining channels. These indices are zero-based and cover the first 20 guided steps.

\subsection{V-Cache}
\label{supp:vcache}

V-Cache approximates IRR's clean-state correction in noisy latent space using the velocity cached from the preceding denoising step. The first guided step uses direct reference insertion; subsequent steps within the correction interval use the cached velocity to correct the latent before model evaluation. The new prediction advances the sampler and updates the cache, avoiding the extra model evaluation used by IRR for correction.

To examine whether this approximation preserves IRR's correction effect, we compare the inferred-noise statistics of the warped-visible and model-generated regions. Direct Insertion, Only IRR, and V-Cache are evaluated on three held-out cases using the same backbone, inputs, target trajectories, initial random seed, noise schedule, and guidance interval.

At denoising step $i$, $x_i$ denotes the latent input to the final model evaluation of that step, and $v_i$ is the corresponding velocity used to update the sampler. For the Flow Matching path $x_\sigma=(1-\sigma)x_0+\sigma\epsilon$, the inferred noise endpoint is
\begin{equation}
\hat{\epsilon}_i = x_i+(1-\sigma_i)v_i.
\label{eq:supp_inferred_noise}
\end{equation}

We compute the regional mean and variance separately for each latent channel, excluding the first conditioning latent frame. Let $p$ index the remaining spatiotemporal positions and $m_p\in[0,1]$ be the warped-visibility mask resampled to the latent resolution. With weights $w_p^{\mathrm{vis}}=m_p$ and $w_p^{\mathrm{gen}}=1-m_p$, the weighted mean and population variance for channel $c$ and region $r\in\{\mathrm{vis},\mathrm{gen}\}$ are
\begin{equation}
\begin{aligned}
\mu_{i,c}^{r}
&= \frac{\sum_p w_p^{r}\hat{\epsilon}_{i,c,p}}
{\sum_p w_p^{r}}, \\
\nu_{i,c}^{r}
&= \frac{\sum_p w_p^{r}
\bigl(\hat{\epsilon}_{i,c,p}-\mu_{i,c}^{r}\bigr)^2}
{\sum_p w_p^{r}}.
\end{aligned}
\label{eq:supp_regional_noise_stats}
\end{equation}

For each case, we compute the absolute differences between the two regions within each channel, then average over the $C$ latent channels:
\begin{equation}
\begin{aligned}
\Delta_{\mu}(i)
&= \frac{1}{C}\sum_{c=1}^{C}
\left|\mu_{i,c}^{\mathrm{vis}}-\mu_{i,c}^{\mathrm{gen}}\right|, \\
\Delta_{\log\mathrm{var}}(i)
&= \frac{1}{C}\sum_{c=1}^{C}
\left|\log
\frac{\nu_{i,c}^{\mathrm{vis}}+\eta}
{\nu_{i,c}^{\mathrm{gen}}+\eta}\right|.
\end{aligned}
\label{eq:supp_noise_gaps}
\end{equation}

Here, $\eta=10^{-12}$ stabilizes the variance ratio, and $\log$ denotes the natural logarithm. Each point in the main-paper curves averages the corresponding gap equally over the three cases at the same denoising step. Smaller gaps indicate closer agreement between the regional noise statistics.

Direct Insertion produces larger gaps, particularly in log-variance, while Only IRR and V-Cache reduce these gaps over much of denoising. These reductions accompany the more coherent structures shown in Fig.~\ref{fig:noise}. V-Cache obtains this improvement using the cached velocity, avoiding IRR's extra model evaluation for correction.

\subsection{Trajectory LoRA Adaptation}
\label{supp:lora_settings}

Trajectory LoRA adapts selected blocks to follow rendered camera guidance while preserving scene content. We construct training pairs from optimized 3D Gaussian representations~\citep{3DGS} of DL3DV scenes~\citep{DL3DV}. For a source view and target camera path, depth-based warping provides a partial reference and its visibility mask, while rendering the scene along the same path provides the target video. Video-VAE encoding gives the trajectory latent $x_{\mathrm{traj}}$ and clean target latent $x_0^{\mathrm{gt}}$. The target video supervises both preservation of visible content and completion of newly exposed regions.

We identify trajectory-sensitive blocks by measuring their contributions to the velocity change learned by a probe LoRA. With the pretrained weights fixed, the probe learns to denoise interpolated trajectory--target states toward the clean target at sampled noise levels. After approximately 400 updates, we evaluate block contributions using 25 trajectory pairs per backbone.

For a fixed validation state $\xi=(x_\sigma,\sigma,\ell)$, let $v_S(\xi)$ be the predicted velocity with probe LoRA enabled in the subset $S\subseteq\mathcal{B}$ of candidate blocks. Enabling all probe adapters changes the frozen backbone's prediction by $d(\xi)=v_{\mathcal{B}}(\xi)-v_{\emptyset}(\xi)$. We measure a subset's contribution along this full response with the normalized utility
\begin{equation}
U_\xi(S)=
\frac{\left\langle v_S(\xi)-v_{\emptyset}(\xi),\,d(\xi)\right\rangle}
{\|d(\xi)\|_2^2}.
\label{eq:supp_shapley_value}
\end{equation}

A block's contribution can depend on which other adapters are active. Shapley scores~\citep{Shapley1953} account for this dependence by averaging the marginal change in $U_\xi$ when a block is enabled across different block combinations. We estimate these scores using random permutations paired with their reverses. Let $S_{\pi,b}$ contain the blocks preceding block $b$ in permutation $\pi$, and let $\Pi$ include the sampled permutations and their reverses. Averaging over $\Pi$ and the validation states $\mathcal{D}$ gives
\begin{equation}
\begin{aligned}
\widehat{\phi}_b
&=\frac{1}{|\mathcal{D}|\,|\Pi|}
\sum_{\xi\in\mathcal{D}}\sum_{\pi\in\Pi}
\Bigl[
U_\xi\bigl(S_{\pi,b}\cup\{b\}\bigr) \\
&\qquad\qquad\qquad\qquad
-U_\xi(S_{\pi,b})
\Bigr].
\end{aligned}
\label{eq:supp_shapley_estimate}
\end{equation}

We select the $K$ highest-scoring self-attention blocks and train LoRA in their query, key, value, and output projections, keeping the pretrained weights fixed. Both backbones use rank $r=128$ and scaling parameter $\alpha=128$, so the update multiplier $\alpha/r$ is one. H3 receives reference conditioning through its Ref2VA branch. Table~\ref{tab:supp_lora_training} lists the default adaptation settings.

\begin{table}[t]
\centering
\caption{Default training settings for Trajectory LoRA. Both backbones first train on recorded guided anchors, then use a mixture of anchors, constructed states, and directly noised targets.}
\label{tab:supp_lora_training}
\footnotesize
\setlength{\tabcolsep}{4pt}
\renewcommand{\arraystretch}{1.12}
\begin{tabular}{@{}lcc@{}}
\toprule
Setting & LongCat & H3 \\
\midrule
Training scenes / pairs & 8 / 50 & 1 / 50 \\
Selected blocks & 36 / 48 & 38 / 50 \\
LoRA rank / $\alpha$ & 128 / 128 & 128 / 128 \\
Optimizer & AdamW & AdamW \\
Learning rate & $2\times10^{-5}$ & $2\times10^{-5}$ \\
Warm-up iterations & 200 & 200 \\
Total training iterations & 1,200 & 1,500 \\
\bottomrule
\end{tabular}
\end{table}

Reference insertion changes the latent states received by the pretrained model, so training begins with states recorded during V-Cache-guided sampling. For each training pair, we save five anchors $(x_a,\sigma_a)$ at approximately $\sigma_a\in\{0.87,0.92,0.93,0.95,0.97\}$. During warm-up, we sample these anchors uniformly, add a small noise-dependent perturbation, and train the selected LoRA parameters to recover the clean target. This initial adaptation supplies the velocity estimates used to construct additional training states in the second stage.

The second stage extends training beyond the saved anchors to cover additional noise levels. In our experiments, 10\% of updates use saved anchors, 25\% use directly noised targets, and 65\% use states constructed from anchors at sampled noise levels. For directly noised targets, $\sigma$ is drawn from $\mathcal{U}(0.08,0.30)$ with probability $0.6$ and from $\mathcal{U}(0.30,0.70)$ otherwise, covering the lower noise levels used to recover fine details. Noise levels for anchor propagation are sampled from the recorded schedule with a multiplicative jitter of up to 5\%.

When a sampled schedule level $\sigma_d$ is close to a saved anchor, the current model predicts $v_a=v_{\theta,\phi}(x_a,\sigma_a,\ell)$ and propagates that anchor with one Euler step:
\begin{equation}
\bar{x}_d=x_a-(\sigma_a-\sigma_d)\operatorname{sg}(v_a),
\label{eq:supp_anchor_propagation}
\end{equation}
where $\operatorname{sg}$ denotes stop-gradient. Within the active guidance interval, we apply reference insertion with the visibility mask and cached-velocity correction to obtain $\tilde{x}_{\sigma_d}$. State construction does not track gradients; the subsequent denoising prediction on the constructed input updates the LoRA parameters.

Both stages use the target velocity defined in the main paper:
\begin{equation}
v_\sigma^{*}=\frac{\tilde{x}_\sigma-x_0^{\mathrm{gt}}}{\sigma}.
\label{eq:supp_lora_target}
\end{equation}

Let $e_\sigma=v_{\theta,\phi}(\tilde{x}_\sigma,\sigma,\ell)-v_\sigma^{*}$, evaluated over the non-conditioning latent elements. The loss covers both visible and newly exposed regions; the first conditioning latent frame remains fixed and is excluded. Both stages use the same squared-error objective as Eq.~\eqref{eq:lora_training_loss} in the main paper:
\begin{equation}
\mathcal{L}_{\mathrm{LoRA}}
=\mathbb{E}\!\left[
\|e_\sigma\|_2^2
\right],
\label{eq:supp_lora_loss}
\end{equation}
where $\theta$ is the frozen backbone, $\phi$ contains the selected LoRA parameters, and $\ell$ denotes the model conditioning.

To study scene diversity, we vary the training scenes among one, two, four, and eight while keeping 50 pairs in total. The update-count study fixes each backbone's training data. Block-selection controls compare equally sized sensitive, random, and low-response subsets, with random-half results averaged over five independent training runs. The training-strategy study fixes inference and compares Standard FM, anchor warm-up alone, guided-state training without warm-up, and the full two-stage strategy, together with a no-LoRA baseline.

For standard sampling, LongCat and H3 use 40 and 50 steps, respectively, and generate videos at approximately $832\times480$ resolution. On each backbone, the trained trajectory LoRA can be combined with the released acceleration LoRA without retraining. Generation times and visual comparisons are provided in Sec.~\ref{supp:acceleration}.

To examine transfer across camera-guidance methods, we reuse the trained LongCat LoRA with Warp-as-History~\citep{WarpAsHistory} and Go-with-the-Flow~\citep{GoWithTheFlow} on the LongCat backbone. Each method provides its own trajectory condition, and the LoRA is applied without further fine-tuning.

\section{Additional Experiments}
\label{supp:additional_experiments}

\subsection{Component Behavior and Motion Selection}
\label{supp:components}

In training-free WorldForge, introducing a rendered reference gives IRR direct control over camera motion, while FLF and DSG help preserve the scene during this change. In Fig.~\ref{fig:supp_components}, FLF reduces appearance distortions from reference fusion, and DSG further corrects the guided prediction using the original prediction. Their combination produces coherent scene content along the requested camera path.

\begin{figure}[!htbp]
\centering
\includegraphics[width=\linewidth]{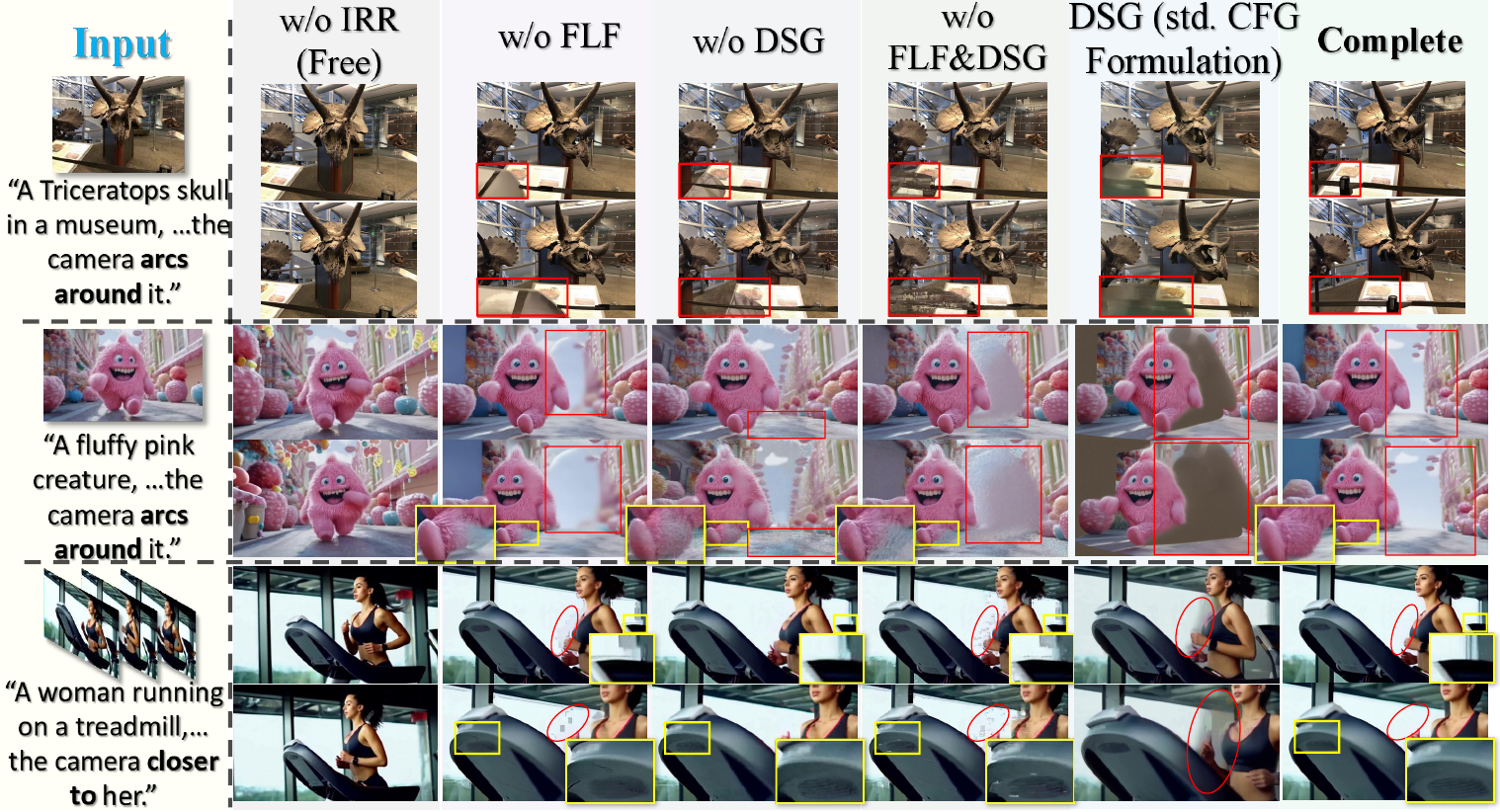}
\caption{Contribution of IRR, FLF, and DSG to training-free camera control. IRR introduces the target camera motion; adding FLF reduces appearance distortions during reference fusion, and DSG further corrects the guided prediction. The complete method preserves coherent scene content along the requested camera path.}
\label{fig:supp_components}
\end{figure}

For training-free control, FLF selects channels by comparing their motion with the rendered reference. Figure~\ref{fig:supp_flf} records how often each channel retains its original prediction across more than 40 static and 30 dynamic scenes. The stable differences in filtering frequency reflect differences in motion relevance: selected channels receive camera guidance, while the remaining channels preserve information from the model's prediction.

\begin{figure}[!htbp]
\centering
\includegraphics[width=0.95\linewidth]{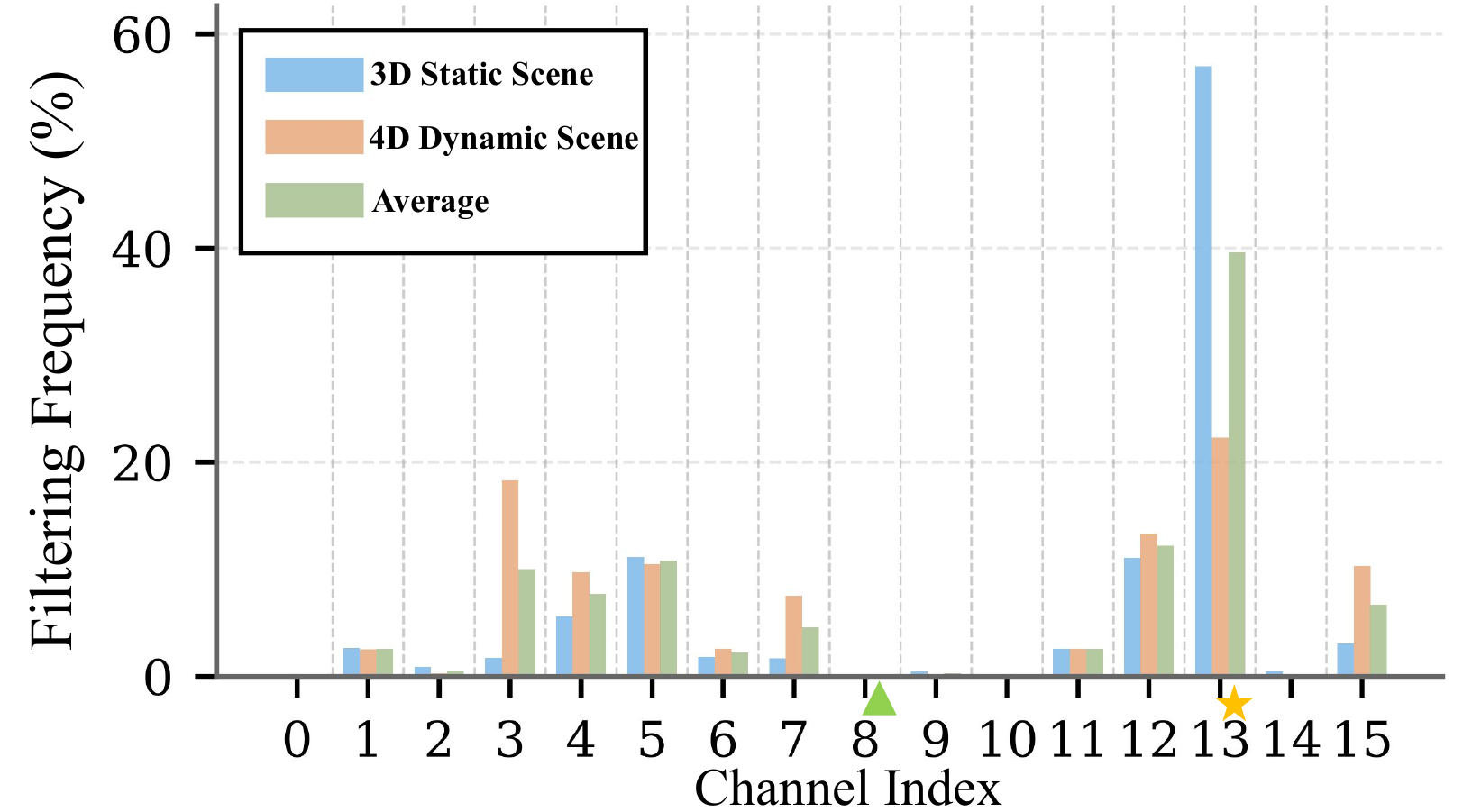}
\caption{Channel filtering frequencies for training-free control over more than 40 static and 30 dynamic scenes. A higher frequency means that a channel more often retains its original prediction. Stable differences across channels support selecting where to inject the camera guidance according to motion relevance.}
\label{fig:supp_flf}
\end{figure}

\subsection{Self-Corrective Guidance}
\label{supp:dsg}

The original and trajectory-guided predictions can point in substantially different directions, making their combination sensitive to the guidance rule. In Fig.~\ref{fig:supp_dsg}, directly applying standard CFG introduces structural artifacts, and removing DSG's adaptive angular weight weakens consistency. The complete DSG uses cosine-based projection and adaptive weighting to combine the predictions, preserving scene structure and fine detail as it follows the rendered guidance.

\begin{figure}[!htbp]
\centering
\includegraphics[width=\linewidth]{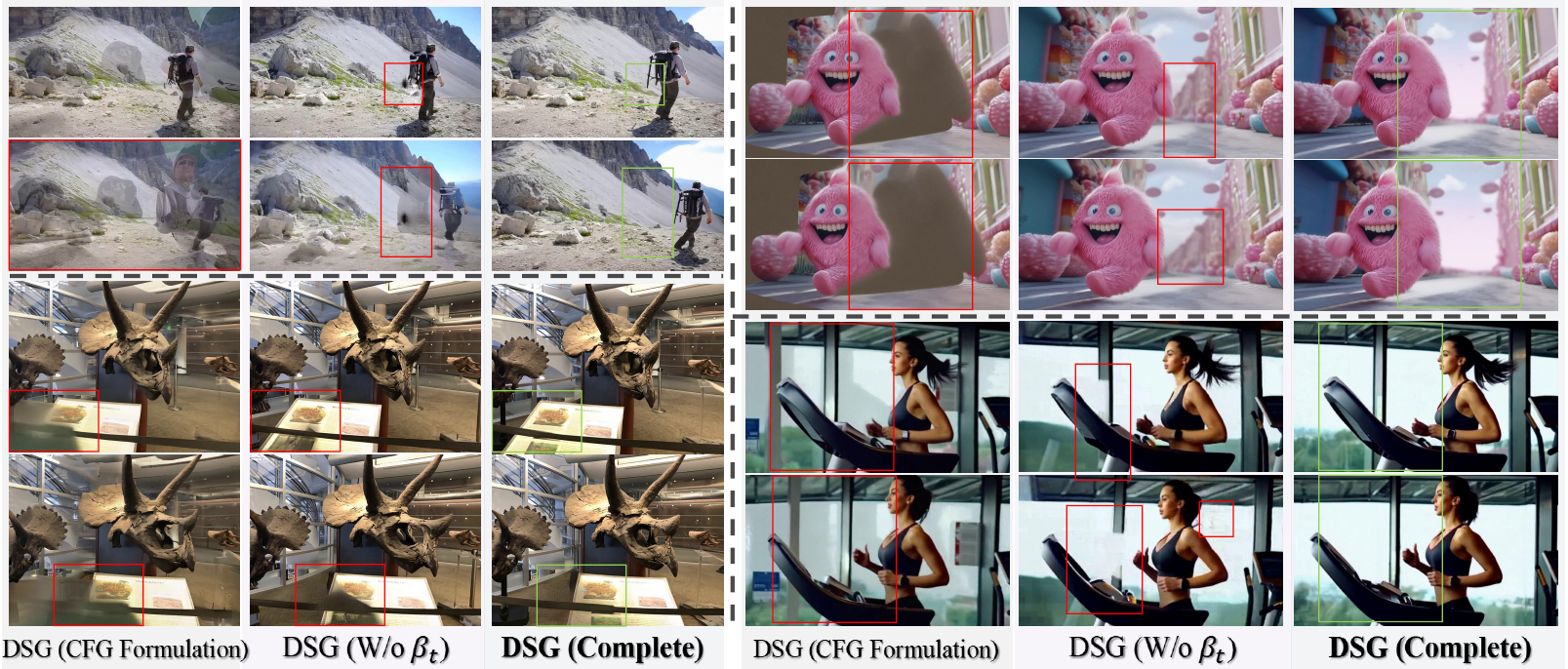}
\caption{Combining the original and trajectory-guided predictions for training-free camera control. Standard CFG introduces structural artifacts, and removing DSG's adaptive angular weight weakens consistency. Cosine-based projection with adaptive weighting preserves scene structure and fine detail in the guided result.}
\label{fig:supp_dsg}
\end{figure}

\subsection{Backbone and Geometry Compatibility}
\label{supp:tf_compatibility}

We apply the same training-free guidance to the U-Net-based SVD~\citep{SVD} and the transformer-based LongCat-Video~\citep{Longcat}. Table~\ref{tab:supp_backbones} compares them with Wan2.1 on more than 40 static scenes. Wan2.1 and LongCat-Video improve both CLIP similarity and camera accuracy over SVD, showing that the shared control mechanism can benefit from stronger video backbones. Figure~\ref{fig:supp_video_backbones} compares guidance methods under a shared SVD backbone and shows the richer scene detail obtained with LongCat-Video.

\begin{table}[!t]
\centering
\caption{Training-free camera control across three video backbones on the static benchmark of more than 40 scenes. All backbones use the same rendered-reference guidance. LongCat-Video leads on CLIP and RPE-T, while Wan2.1 leads on ATE and RPE-R; both improve all reported metrics over SVD. Bold and underlining mark the best and second-best results.}
\label{tab:supp_backbones}
\footnotesize
\setlength{\tabcolsep}{3pt}
\begin{tabular}{lcccc}
\toprule
Backbone & CLIP$\uparrow$ & ATE$\downarrow$ & RPE-T$\downarrow$ & RPE-R$\downarrow$ \\
\midrule
SVD & 0.910 & 0.265 & 0.316 & 0.444 \\
Wan2.1 & \underline{0.948} & \textbf{0.077} & \underline{0.086} & \textbf{0.221} \\
LongCat-Video & \textbf{0.949} & \underline{0.095} & \textbf{0.076} & \underline{0.230} \\
\bottomrule
\end{tabular}

\end{table}
\begin{figure}[!htbp]
\centering
\includegraphics[width=\linewidth]{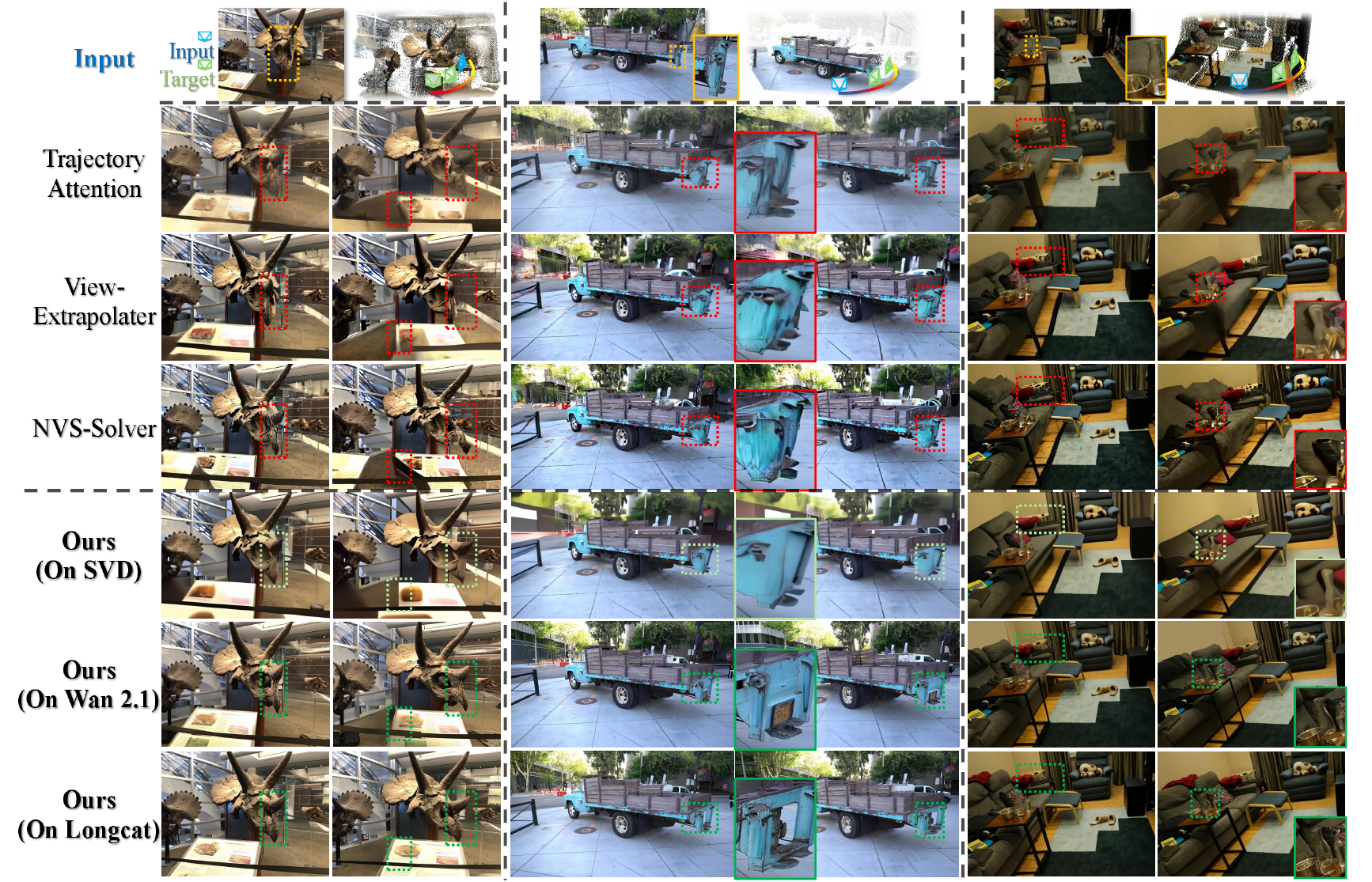}
\caption{Training-free camera control across video backbones. Comparisons under a shared SVD backbone show the effect of the guidance method. WorldForge also produces richer scene detail with LongCat-Video, using the same inference-time guidance.}
\label{fig:supp_video_backbones}
\end{figure}
\begin{table*}[!t]
  \centering
  \caption{Training resources and inference efficiency for camera-controlled generation. Our trajectory adapters learn from 50 videos with 1,200 or 1,500 updates. Combining them with released acceleration LoRAs gives speedups of $4.09\times$ on LongCat and $5.09\times$ on H3. Times are mean seconds per generated video over 200 test cases for each method, estimated for a single A100 from local runs with native video configurations. Parameter counts refer to the backbones. Training updates exclude backbone pretraining; a dash denotes an unreported total update count.}
  \label{tab:supp_efficiency}
  \label{tab:supp_acceleration}
  \fontsize{8.5}{10}\selectfont
  \setlength{\tabcolsep}{2pt}
  \renewcommand{\arraystretch}{1.15}
  \begin{tabular*}{\textwidth}{@{\extracolsep{\fill}}llrlrr@{}}
  \toprule
  Method & Backbone & \shortstack{Backbone\\params.} & Training data & \shortstack{Training\\updates} & \shortstack{Inference\\time (s)} \\
  \midrule
  UniWorld-View~\citep{uniworld} & Wan2.1-VACE & 14B & 100K triplets + 100K pairs & 20,000 & 124.5 \\
  TrajectoryCrafter~\citep{TrajCrafter} & CogVideoX-Fun & 5B & 120K multi-view + 60K pairs & 15,000 & 169.0 \\
  SymphoMotion~\citep{symphomotion} & Wan2.1-I2V & 14B & 25K videos & --- & 442.9 \\
  Warp-as-History~\citep{WarpAsHistory} & Helios & 14B & 1 scene / 4 clips & 1,000 & 39.3 \\
  \midrule
  Ours (LongCat) & LongCat-Video & 13.6B & 8 scenes / 50 videos & 1,200 & 274.2 \\
  Ours (LongCat+acceleration LoRA) & LongCat-Video & 13.6B & 8 scenes / 50 videos & 1,200 & 67.0 \\
  Ours (H3) & MiniMax-H3 & 33B & 1 scene / 50 videos & 1,500 & 707.3 \\
  Ours (H3+acceleration LoRA) & MiniMax-H3 & 33B & 1 scene / 50 videos & 1,500 & 139.0 \\
  \bottomrule
  \end{tabular*}
  \end{table*}

For training-free guidance, FLF uses optical flow to compare predicted motion with the rendered reference. Replacing the lightweight Farneb\"ack estimator~\citep{OpticaFlow} with RAFT allows us to examine whether this selection remains useful with a learned flow model. Figure~\ref{fig:supp_flow_estimators}(a) shows that FLF reduces rendering artifacts and improves generation quality with both estimators.

We also construct warped references with VGGT, UniDepth, Mega-SaM, and DepthCrafter to examine the effect of depth estimation (Fig.~\ref{fig:supp_depth}(b)). Although the estimated geometry differs, the generated views maintain recognizable layouts and plausible surfaces across these reference inputs.

\begin{figure}[!htbp]
\centering
\includegraphics[width=\linewidth]{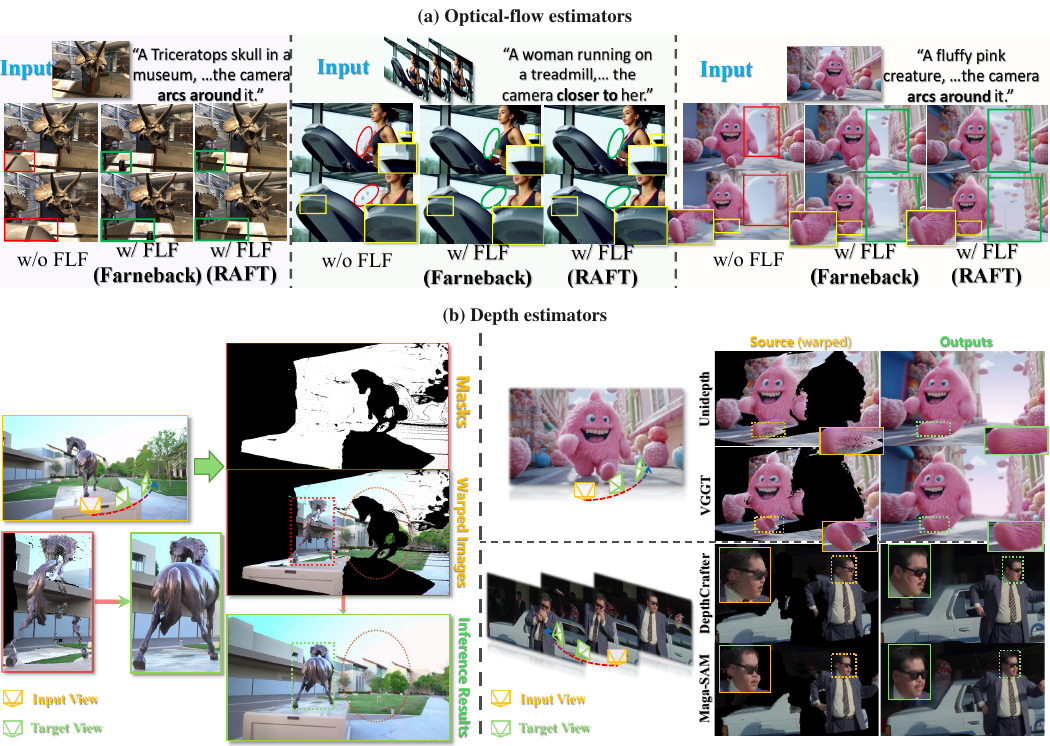}
\caption{Compatibility of training-free guidance with motion and geometry estimators. (a) FLF reduces rendering artifacts with both Farneb\"ack and RAFT optical flow. (b) Warped references from VGGT, UniDepth, Mega-SaM, and DepthCrafter support coherent scene layouts and plausible surfaces despite differences in their estimated geometry.}
\label{fig:supp_flow_estimators}
\label{fig:supp_depth}
\end{figure}

\subsection{Training Resources and Inference Efficiency}
\label{supp:efficiency}
\label{supp:acceleration}

Trajectory LoRA learns camera control from 50 videos of 25--93 frames while keeping the pretrained backbone fixed. LongCat uses eight scenes and 1,200 updates, and H3 uses one scene and 1,500 updates. Table~\ref{tab:supp_efficiency} compares these training resources and the resulting generation time with the other methods. Warp-as-History also uses a compact source set, deriving four training clips from one source video and expanding them through online clip and direction augmentation.

For each backbone, the trained trajectory LoRA can be reused with a released acceleration LoRA, reducing sampling cost without further trajectory training. LongCat~\citep{Longcat} reduces its sampling steps from 40 to 16; H3 uses MiniMax-H3 Turbo LoRA to reduce sampling from 50 to eight. Visual results are shown in Fig.~\ref{fig:supp_acceleration}.

\begin{figure}[!htbp]
\centering
\includegraphics[width=\linewidth]{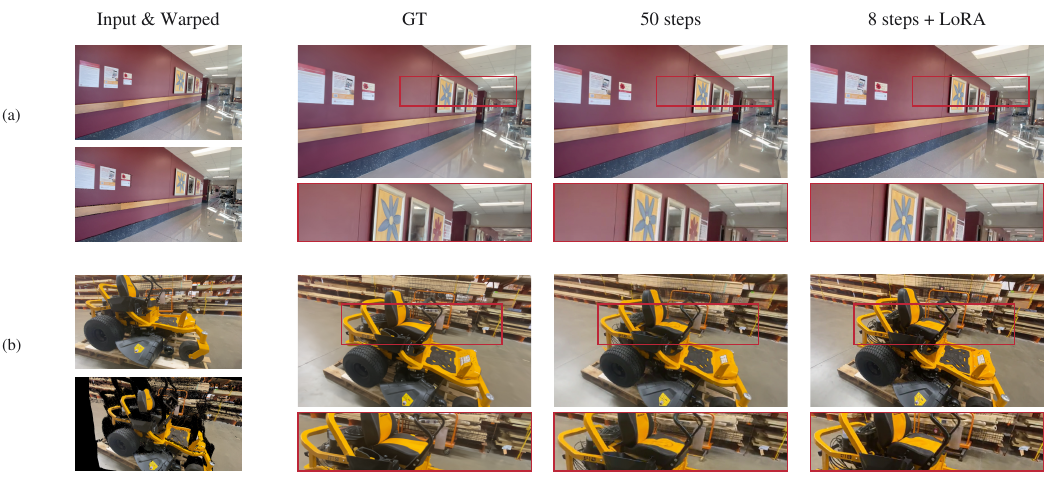}
\caption{H3 generation with 50-step and eight-step sampling using the same trajectory LoRA. The accelerated results retain scene layout and object geometry in both examples. Enlargements compare corresponding details.}
\label{fig:supp_acceleration}
\end{figure}

\subsection{Challenging Scenes and Control Limits}
\label{supp:challenging}

\begin{figure}[!htbp]
  \centering
  \includegraphics[width=\linewidth]{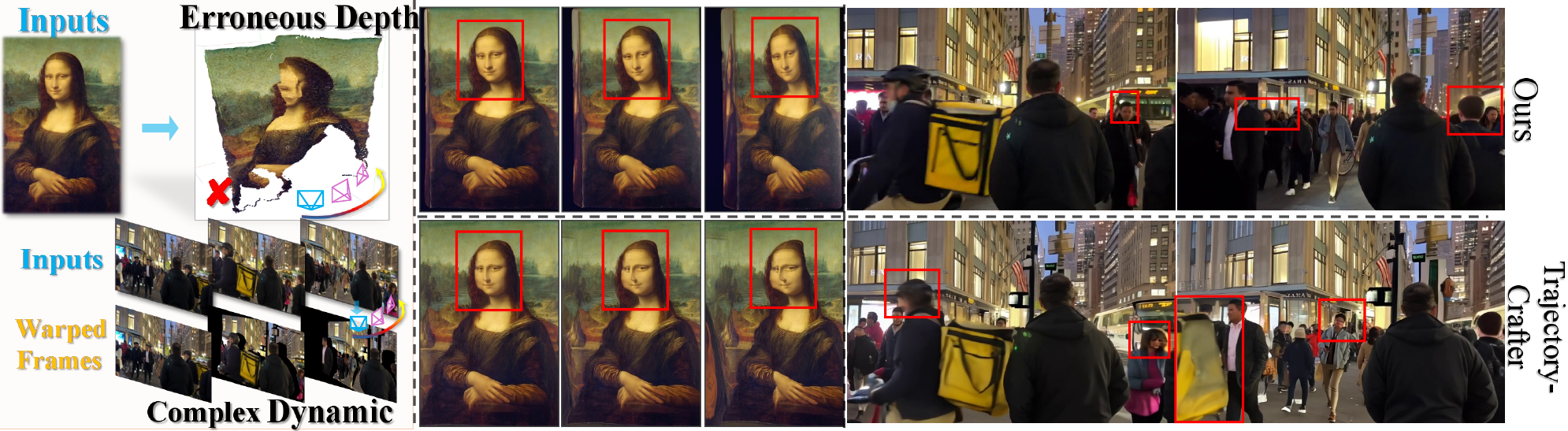}
  \caption{Camera-control limits under erroneous depth and complex scene dynamics. Distorted face geometry and inconsistencies around moving people show how reference errors remain visible in the generated views. Rows compare training-free WorldForge (top) with TrajectoryCrafter (bottom).}
  \label{fig:supp_control_limits}
  \end{figure}
Fast motion, non-rigid motion, and heavy occlusion make reference fusion more difficult. Figure~\ref{fig:supp_challenging_flf} compares generation with and without FLF under these conditions. Selecting the channels for fusion reduces the marked distortions in objects, water, and scene boundaries while retaining the requested viewpoint changes.

\begin{figure}[!htbp]
\centering
\includegraphics[width=\linewidth]{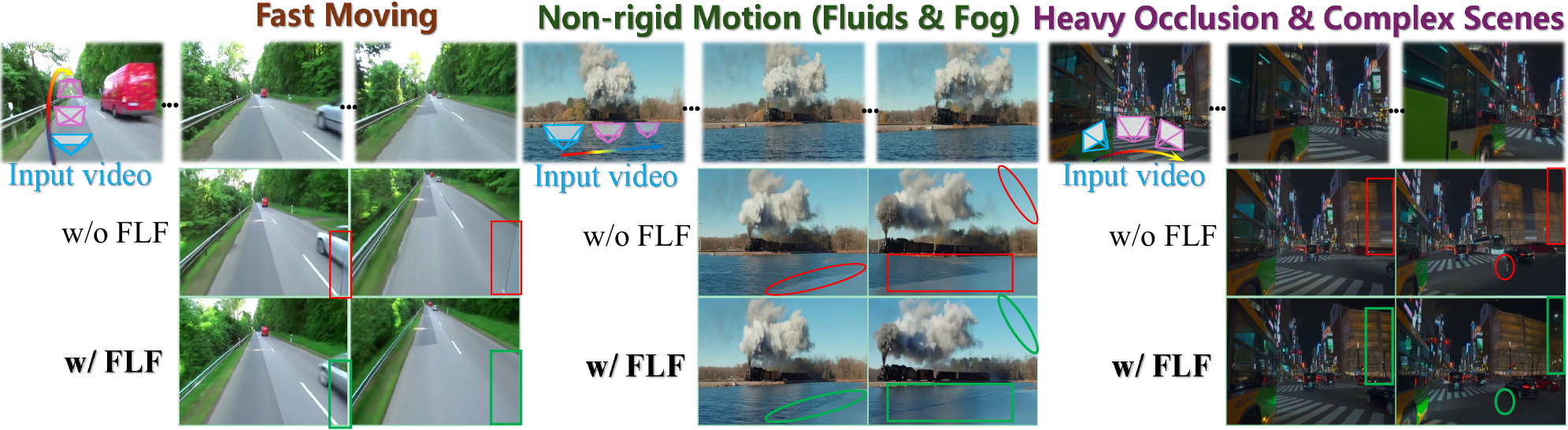}
\caption{FLF in training-free generation under fast motion, non-rigid motion, and heavy occlusion. Comparing w/o FLF with w/ FLF shows that selecting channels for reference fusion reduces distortions in objects, water, and scene boundaries while retaining the requested viewpoint changes.}
\label{fig:supp_challenging_flf}
\end{figure}

\begin{figure}[!htbp]
  \centering
  \includegraphics[width=\linewidth]{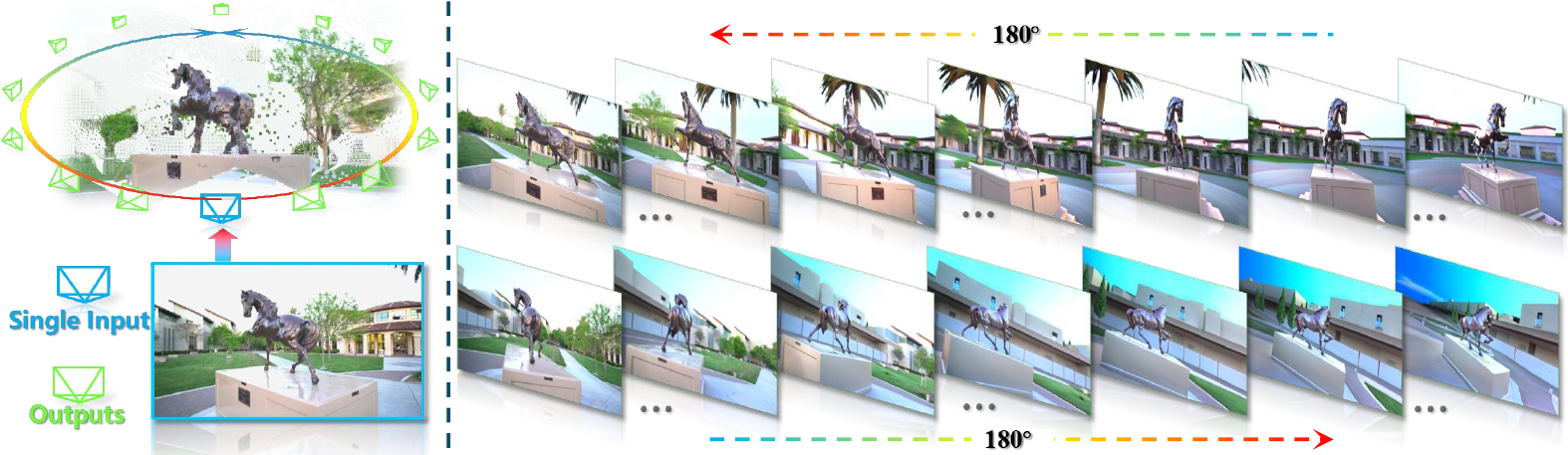}
  \caption{Training-free generation of an object-centric $360^\circ$ orbit from one input image. The final frame of each clip initializes the next segment, extending camera control beyond the original field of view while retaining recognizable scene content.}
  \label{fig:supp_orbit}
  \end{figure}

\begin{figure*}[!t]
  \centering
  \includegraphics[width=0.9\linewidth]{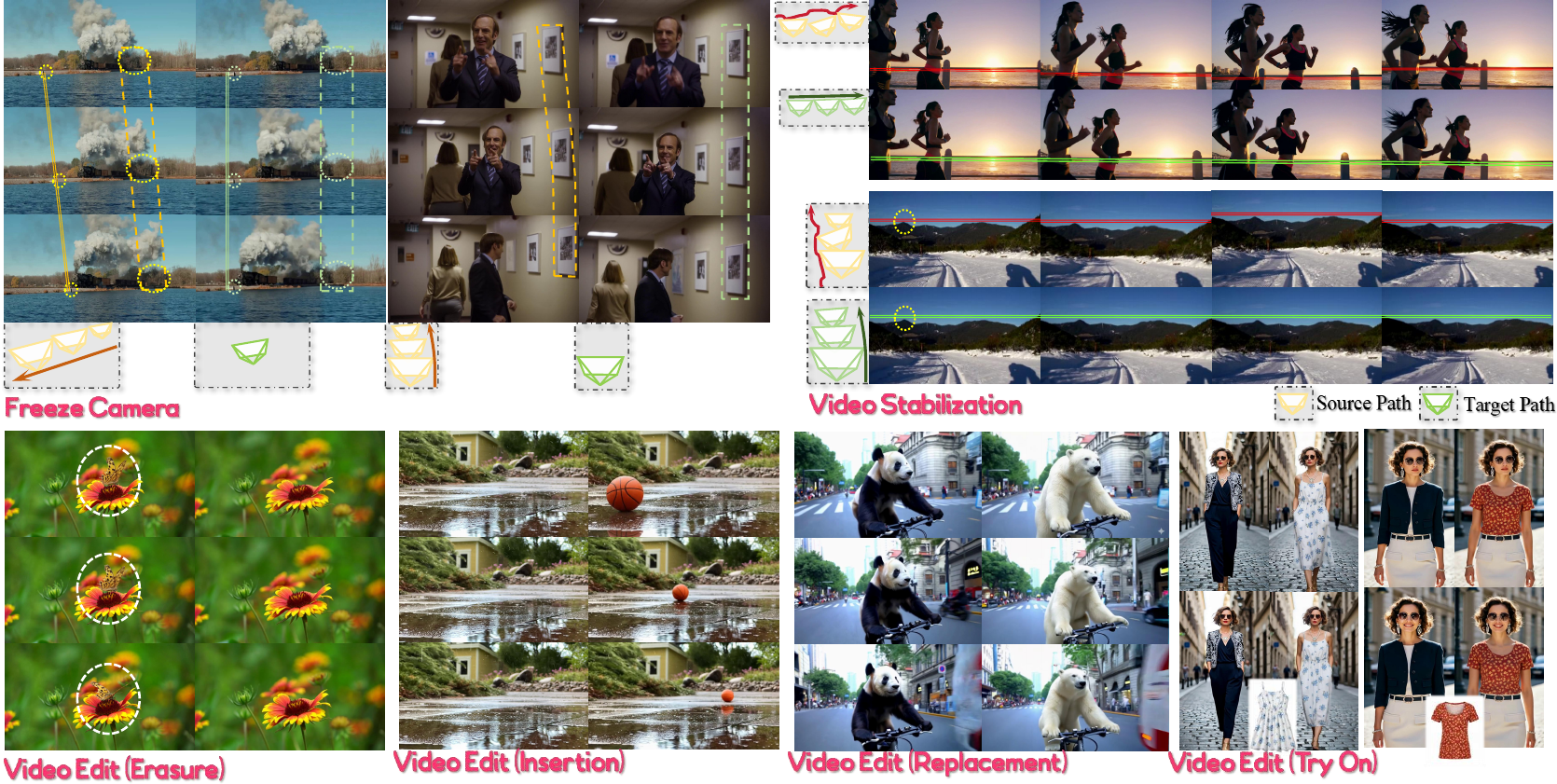}
  \caption{Camera and content editing with training-free WorldForge. Rendered references support camera freezing and stabilization while retaining scene content. An edited first frame and region masks propagate appearance changes through the video.}
  \label{fig:supp_editing}
  \end{figure*}

Severe depth errors can give the rendered reference incorrect geometry, while complex scene dynamics can make its guidance inconsistent across frames. Figure~\ref{fig:supp_control_limits} shows these errors in the outputs of training-free WorldForge and TrajectoryCrafter, including distorted faces and inconsistencies around moving people.
\subsection{Wide-Range Control and Editing}
\label{supp:tf_applications}

\begin{figure*}[!t]
  \centering
  \includegraphics[width=0.86\linewidth]{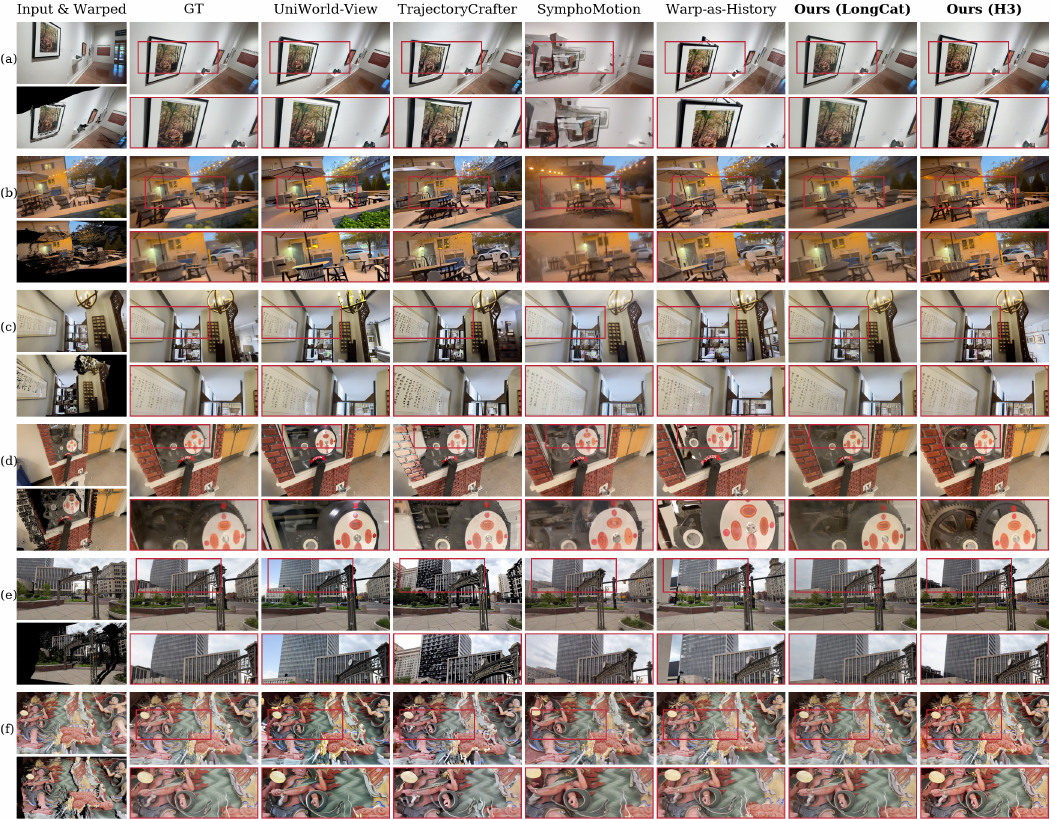}
  \caption{Additional camera-controlled generation on six held-out DL3DV cases beyond the main figure. Our LongCat and H3 results preserve scene layout and local appearance, while several baseline outputs exhibit distorted boundaries, texture loss, or rendering remnants. LongCat uses the main-table model; H3 uses the eight-step accelerated sampling evaluated in Sec.~\ref{supp:acceleration}. Inputs and warped references appear on the left, and red boxes identify corresponding enlarged regions.}
  \label{fig:supp_gallery}
  \end{figure*}

To follow paths with large camera rotations, we generate successive clips and use the final frame of each clip to initialize the next. Figure~\ref{fig:supp_orbit} shows an object-centric $360^\circ$ orbit generated from one input image. The clips retain recognizable scene content as the camera moves beyond the original field of view.

The same rendered-reference guidance also supports camera freezing and stabilization (Fig.~\ref{fig:supp_editing}). For content editing, we segment the target region in each frame and edit its appearance in the first frame. This edited frame and the region masks guide the change through the video, while the rendered reference specifies the viewpoint. For an object absent from the source video, a bounding box specifies where it should be added.

\section{Additional Visual Comparisons}
\label{supp:gallery}

Figure~\ref{fig:supp_gallery} adds six held-out DL3DV cases to the learned-control comparison. Indoor layouts, outdoor structures, and finely textured objects allow us to examine how viewpoint changes affect structural boundaries, repeated patterns, and surface detail. Our LongCat and H3 results preserve scene layout and local appearance, while several baseline outputs contain distorted boundaries, texture loss, or rendering remnants.

\FloatBarrier
\clearpage
\endgroup


\begin{thebibliography}{99}
\providecommand{\url}[1]{#1}
\csname url@samestyle\endcsname
\providecommand{\newblock}{\relax}
\providecommand{\bibinfo}[2]{#2}
\providecommand{\BIBentrySTDinterwordspacing}{\spaceskip=0pt\relax}
\providecommand{\BIBentryALTinterwordstretchfactor}{4}
\providecommand{\BIBentryALTinterwordspacing}{\spaceskip=\fontdimen2\font plus
\BIBentryALTinterwordstretchfactor\fontdimen3\font minus \fontdimen4\font\relax}
\providecommand{\BIBforeignlanguage}[2]{{\expandafter\ifx\csname l@#1\endcsname\relax
\typeout{** WARNING: IEEEtran.bst: No hyphenation pattern has been}\typeout{** loaded for the language `#1'. Using the pattern for}\typeout{** the default language instead.}\else
\language=\csname l@#1\endcsname
\fi
#2}}
\providecommand{\BIBdecl}{\relax}
\BIBdecl

\bibitem{SVD}
A.~Blattmann \emph{et~al.}, ``Stable video diffusion: Scaling latent video diffusion models to large datasets,'' \emph{arXiv preprint arXiv:2311.15127}, 2023.

\bibitem{Wan21}
{Team Wan} \emph{et~al.}, ``Wan: Open and advanced large-scale video generative models,'' \emph{arXiv preprint arXiv:2503.20314}, 2025.

\bibitem{CogVideoX}
Z.~Yang \emph{et~al.}, ``CogVideoX: Text-to-video diffusion models with an expert transformer,'' in \emph{International Conference on Learning Representations (ICLR)}, 2025.

\bibitem{ViewCrafter}
W.~Yu \emph{et~al.}, ``ViewCrafter: Taming video diffusion models for high-fidelity novel view synthesis,'' \emph{IEEE Transactions on Pattern Analysis and Machine Intelligence}, early access, pp. 1--18, 2025, doi: 10.1109/TPAMI.2025.3613256.

\bibitem{TrajCrafter}
M.~Yu, W.~Hu, J.~Xing, and Y.~Shan, ``TrajectoryCrafter: Redirecting camera trajectory for monocular videos via diffusion models,'' in \emph{Proceedings of the IEEE/CVF International Conference on Computer Vision (ICCV)}, 2025, pp. 100--111.

\bibitem{NVSSolver}
M.~You, Z.~Zhu, H.~Liu, and J.~Hou, ``NVS-Solver: Video diffusion model as zero-shot novel view synthesizer,'' in \emph{International Conference on Learning Representations (ICLR)}, 2025.

\bibitem{VideoScene}
H.~Wang, F.~Liu, J.~Chi, and Y.~Duan, ``VideoScene: Distilling video diffusion model to generate 3D scenes in one step,'' in \emph{Proceedings of the IEEE/CVF Conference on Computer Vision and Pattern Recognition (CVPR)}, 2025, pp. 16\,475--16\,485.

\bibitem{Free4D}
T.~Liu \emph{et~al.}, ``Free4D: Tuning-free 4D scene generation with spatial-temporal consistency,'' in \emph{Proceedings of the IEEE/CVF International Conference on Computer Vision (ICCV)}, 2025, pp. 25\,571--25\,582.

\bibitem{Genie}
J.~Bruce \emph{et~al.}, ``Genie: Generative interactive environments,'' in \emph{Proceedings of the 41st International Conference on Machine Learning}, ser. Proceedings of Machine Learning Research, vol.~235, 2024, pp. 4603--4623.

\bibitem{NavigationWM}
A.~Bar, G.~Zhou, D.~Tran, T.~Darrell, and Y.~LeCun, ``Navigation world models,'' in \emph{Proceedings of the IEEE/CVF Conference on Computer Vision and Pattern Recognition (CVPR)}, 2025, pp. 15\,791--15\,801.

\bibitem{4DSpatialSurvey}
Y.~Cao \emph{et~al.}, ``Reconstructing 4D spatial intelligence: A survey,'' \emph{arXiv preprint arXiv:2507.21045}, 2025.

\bibitem{MotionCtrl}
Z.~Wang \emph{et~al.}, ``MotionCtrl: A unified and flexible motion controller for video generation,'' in \emph{ACM SIGGRAPH 2024 Conference Papers}, 2024, pp. 1--11.

\bibitem{CameraCtrl}
H.~He \emph{et~al.}, ``CameraCtrl: Enabling camera control for video diffusion models,'' in \emph{International Conference on Learning Representations (ICLR)}, 2025.

\bibitem{ReCamMaster}
J.~Bai \emph{et~al.}, ``ReCamMaster: Camera-controlled generative rendering from a single video,'' in \emph{Proceedings of the IEEE/CVF International Conference on Computer Vision (ICCV)}, 2025, pp. 14\,834--14\,844.

\bibitem{TrajectoryAttention}
Z.~Xiao \emph{et~al.}, ``Trajectory attention for fine-grained video motion control,'' in \emph{International Conference on Learning Representations (ICLR)}, 2025.

\bibitem{GoWithTheFlow}
R.~Burgert \emph{et~al.}, ``Go-with-the-Flow: Motion-controllable video diffusion models using real-time warped noise,'' in \emph{Proceedings of the IEEE/CVF Conference on Computer Vision and Pattern Recognition (CVPR)}, 2025, pp. 13--23.

\bibitem{ViewExtrapolator}
K.~Liu, L.~Shao, and S.~Lu, ``Novel view extrapolation with video diffusion priors,'' \emph{arXiv preprint arXiv:2411.14208}, 2024.

\bibitem{WarpAsHistory}
Y.~Wang and T.~He, ``Warp-as-History: Generalizable camera-controlled video generation from one training video,'' \emph{arXiv preprint arXiv:2605.15182}, 2026.

\bibitem{CFG}
J.~Ho and T.~Salimans, ``Classifier-free diffusion guidance,'' in \emph{NeurIPS Workshop on Deep Generative Models and Downstream Applications}, 2021.

\bibitem{MOFT}
Z.~Xiao, Y.~Zhou, S.~Yang, and X.~Pan, ``Video diffusion models are training-free motion interpreter and controller,'' in \emph{Advances in Neural Information Processing Systems}, vol.~37, 2024, pp. 76\,115--76\,138.

\bibitem{worldforge}
C.~Song, Y.~Yang, T.~Zhao, R.~Li, and C.~Zhang, ``Taming video models for 3D and 4D generation via zero-shot camera control,'' in \emph{Proceedings of the IEEE/CVF Conference on Computer Vision and Pattern Recognition (CVPR)}, 2026, pp. 40\,352--40\,363.

\bibitem{HunyuanVideo}
W.~Kong \emph{et~al.}, ``HunyuanVideo: A systematic framework for large video generative models,'' \emph{arXiv preprint arXiv:2412.03603}, 2024.

\bibitem{AC3D}
S.~Bahmani \emph{et~al.}, ``AC3D: Analyzing and improving 3D camera control in video diffusion transformers,'' in \emph{Proceedings of the IEEE/CVF Conference on Computer Vision and Pattern Recognition (CVPR)}, 2025, pp. 22\,875--22\,889.

\bibitem{VD3D}
S.~Bahmani \emph{et~al.}, ``VD3D: Taming large video diffusion transformers for 3D camera control,'' in \emph{International Conference on Learning Representations (ICLR)}, 2025.

\bibitem{GCD}
B.~Van~Hoorick \emph{et~al.}, ``Generative camera dolly: Extreme monocular dynamic novel view synthesis,'' in \emph{European Conference on Computer Vision (ECCV)}, 2024, pp. 313--331.

\bibitem{LingBotWorld}
{Robbyant Team} \emph{et~al.}, ``Advancing open-source world models,'' \emph{arXiv preprint arXiv:2601.20540}, 2026.

\bibitem{InSpatioWorld}
{InSpatio Team} \emph{et~al.}, ``INSPATIO-WORLD: A real-time 4D world simulator via spatiotemporal autoregressive modeling,'' \emph{arXiv preprint arXiv:2604.07209}, 2026.

\bibitem{HYWorld2}
{Team HY-World} \emph{et~al.}, ``HY-World 2.0: A multi-modal world model for reconstructing, generating, and simulating 3D worlds,'' \emph{arXiv preprint arXiv:2604.14268}, 2026.

\bibitem{WorldPlay}
W.~Sun \emph{et~al.}, ``WorldPlay: Towards long-term geometric consistency for real-time interactive world modeling,'' in \emph{Proceedings of the 43rd International Conference on Machine Learning (ICML)}, ser. Proceedings of Machine Learning Research, vol.~306, 2026, pp. 117\,501--117\,523.

\bibitem{MatrixGame3}
Z.~Wang \emph{et~al.}, ``Matrix-Game 3.0: Real-time and streaming interactive world model with long-horizon memory,'' \emph{arXiv preprint arXiv:2604.08995}, 2026.

\bibitem{NeRF}
B.~Mildenhall, P.~P. Srinivasan, M.~Tancik, J.~T. Barron, R.~Ramamoorthi, and R.~Ng, ``NeRF: Representing scenes as neural radiance fields for view synthesis,'' in \emph{European Conference on Computer Vision (ECCV)}, 2020, pp. 405--421.

\bibitem{3DGS}
B.~Kerbl, G.~Kopanas, T.~Leimkuehler, and G.~Drettakis, ``3D Gaussian splatting for real-time radiance field rendering,'' \emph{ACM Transactions on Graphics (TOG)}, vol.~42, no.~4, Art. no.~139, pp. 1--14, 2023.

\bibitem{MVSNet}
Y.~Yao, Z.~Luo, S.~Li, T.~Fang, and L.~Quan, ``MVSNet: Depth inference for unstructured multi-view stereo,'' in \emph{European Conference on Computer Vision (ECCV)}, 2018, pp. 767--783.

\bibitem{Mipsplatting}
Z.~Yu, A.~Chen, B.~Huang, T.~Sattler, and A.~Geiger, ``Mip-Splatting: Alias-free 3D Gaussian splatting,'' in \emph{Proceedings of the IEEE/CVF Conference on Computer Vision and Pattern Recognition (CVPR)}, 2024, pp. 19\,447--19\,456.

\bibitem{InstantNGP}
T.~M{\"u}ller, A.~Evans, C.~Schied, and A.~Keller, ``Instant neural graphics primitives with a multiresolution hash encoding,'' \emph{ACM Transactions on Graphics (TOG)}, vol.~41, no.~4, Art. no.~102, pp. 1--15, 2022.

\bibitem{Fewarnet}
C.~Song, S.~Wang, J.~Wei, and Y.~Zhao, ``FewarNet: An efficient few-shot view synthesis network based on trend regularization,'' \emph{IEEE Transactions on Circuits and Systems for Video Technology (TCSVT)}, vol.~34, no.~10, pp. 9264--9280, 2024.

\bibitem{PDFSV}
C.~Song, S.~Wang, J.~Wei, Y.~Zhao, and R.~Zhang, ``Wide-baseline view synthesis for light-field display based on plane-depth-fused sweep volume,'' \emph{Displays}, vol.~79, Art. no.~102503, 2023.

\bibitem{PERF}
G.~Wang, P.~Wang, Z.~Chen, W.~Wang, C.~C. Loy, and Z.~Liu, ``PERF: Panoramic neural radiance field from a single panorama,'' \emph{IEEE Transactions on Pattern Analysis and Machine Intelligence}, vol.~46, no.~10, pp. 6905--6918, 2024, doi: 10.1109/TPAMI.2024.3387307.

\bibitem{DreamFusion}
B.~Poole, A.~Jain, J.~T. Barron, and B.~Mildenhall, ``DreamFusion: Text-to-3D using 2D diffusion,'' in \emph{International Conference on Learning Representations (ICLR)}, 2023.

\bibitem{ASD}
M.~Wei, J.~Zhou, J.~Sun, and X.~Zhang, ``Adversarial score distillation: When score distillation meets GAN,'' in \emph{Proceedings of the IEEE/CVF Conference on Computer Vision and Pattern Recognition (CVPR)}, 2024, pp. 8131--8141.

\bibitem{Trellis}
J.~Xiang \emph{et~al.}, ``Structured 3D latents for scalable and versatile 3D generation,'' in \emph{Proceedings of the IEEE/CVF Conference on Computer Vision and Pattern Recognition (CVPR)}, 2025, pp. 21\,469--21\,480.

\bibitem{GeometryAwareSDS}
M.-S. Kwak, D.~Ahn, I.~H. Kim, J.-H. Kim, and S.~Kim, ``Geometry-aware score distillation via 3D consistent noising and gradient consistency modeling,'' \emph{arXiv preprint arXiv:2406.16695}, 2024.

\bibitem{DreamComposerPlusPlus}
Y.~Yang \emph{et~al.}, ``DreamComposer++: Empowering diffusion models with multi-view conditions for 3D content generation,'' \emph{IEEE Transactions on Pattern Analysis and Machine Intelligence}, early access, pp. 1--18, 2025, doi: 10.1109/TPAMI.2025.3568190.

\bibitem{SyncamMaster}
J.~Bai \emph{et~al.}, ``SynCamMaster: Synchronizing multi-camera video generation from diverse viewpoints,'' in \emph{International Conference on Learning Representations (ICLR)}, 2025.

\bibitem{FYC}
Y.~Ma \emph{et~al.}, ``EasyCreator: Empowering 4D creation through video inpainting,'' in \emph{International Conference on Learning Representations (ICLR)}, 2026.

\bibitem{VUD}
F.~Tian, T.~Ding, J.~Luo, H.~Min, and R.~Vidal, ``Voyaging into perpetual dynamic scenes from a single view,'' in \emph{Proceedings of the IEEE/CVF International Conference on Computer Vision (ICCV)}, 2025, pp. 7698--7708.

\bibitem{LatentReframe}
Z.~Zhou, J.~An, and J.~Luo, ``Latent-Reframe: Enabling camera control for video diffusion models without training,'' in \emph{Proceedings of the IEEE/CVF International Conference on Computer Vision (ICCV)}, 2025, pp. 12\,779--12\,789.

\bibitem{WorldInWorld}
C.~Song, Y.~Yang, and C.~Zhang, ``World in world: Explore the world with world models,'' \emph{arXiv preprint arXiv:2609.11548}, 2026.

\bibitem{See3D}
B.~Ma \emph{et~al.}, ``You see it, you got it: Learning 3D creation on pose-free videos at scale,'' in \emph{Proceedings of the IEEE/CVF Conference on Computer Vision and Pattern Recognition (CVPR)}, 2025, pp. 2016--2029.

\bibitem{Voyager}
T.~Huang \emph{et~al.}, ``Voyager: Long-range and world-consistent video diffusion for explorable 3D scene generation,'' \emph{ACM Transactions on Graphics (TOG)}, vol.~44, no.~6, Art. no.~245, pp. 1--15, 2025.

\bibitem{TrajMaster3D}
X.~Fu \emph{et~al.}, ``3DTrajMaster: Mastering 3D trajectory for multi-entity motion in video generation,'' in \emph{International Conference on Learning Representations (ICLR)}, 2025.

\bibitem{BadGuide}
T.~Karras, M.~Aittala, T.~Kynk{\"a}{\"a}nniemi, J.~Lehtinen, T.~Aila, and S.~Laine, ``Guiding a diffusion model with a bad version of itself,'' \emph{Advances in Neural Information Processing Systems (NeurIPS)}, vol.~37, pp. 52\,996--53\,021, 2024.

\bibitem{STG}
J.~Hyung, K.~Kim, S.~Hong, M.-J. Kim, and J.~Choo, ``Spatiotemporal skip guidance for enhanced video diffusion sampling,'' in \emph{Proceedings of the IEEE/CVF Conference on Computer Vision and Pattern Recognition (CVPR)}, 2025, pp. 11\,006--11\,015.

\bibitem{APG}
S.~Sadat, O.~Hilliges, and R.~M. Weber, ``Eliminating oversaturation and artifacts of high guidance scales in diffusion models,'' in \emph{International Conference on Learning Representations (ICLR)}, 2025.

\bibitem{FlowDirector}
G.~Li, Y.~Yang, C.~Song, X.~Liu, and C.~Zhang, ``FlowDirector: Training-free flow steering for precise text-to-video editing,'' in \emph{Proceedings of the IEEE/CVF Conference on Computer Vision and Pattern Recognition (CVPR)}, 2026, pp. 7805--7815.

\bibitem{RestartSampling}
Y.~Xu, M.~Deng, X.~Cheng, Y.~Tian, Z.~Liu, and T.~Jaakkola, ``Restart sampling for improving generative processes,'' \emph{Advances in Neural Information Processing Systems (NeurIPS)}, vol.~36, pp. 76\,806--76\,838, 2023.

\bibitem{Repaint}
A.~Lugmayr, M.~Danelljan, A.~Romero, F.~Yu, R.~Timofte, and L.~Van~Gool, ``RePaint: Inpainting using denoising diffusion probabilistic models,'' in \emph{Proceedings of the IEEE/CVF Conference on Computer Vision and Pattern Recognition (CVPR)}, 2022, pp. 11\,461--11\,471.

\bibitem{zigzag}
L.~Bai \emph{et~al.}, ``Zigzag diffusion sampling: Diffusion models can self-improve via self-reflection,'' in \emph{International Conference on Learning Representations (ICLR)}, 2025.

\bibitem{D2T}
X.~Cai \emph{et~al.}, ``Dust to tower: Prior-driven coarse-to-fine photo-realistic scene reconstruction from sparse uncalibrated images,'' \emph{IEEE Transactions on Visualization and Computer Graphics (TVCG)}, vol.~32, no.~7, pp. 6129--6143, 2026.

\bibitem{VistaDream}
H.~Wang, Y.~Liu, Z.~Liu, W.~Wang, Z.~Dong, and B.~Yang, ``{VistaDream}: Sampling multiview consistent images for single-view scene reconstruction,'' in \emph{Proceedings of the IEEE/CVF International Conference on Computer Vision (ICCV)}, 2025, pp. 26\,772--26\,782.

\bibitem{ControlNet}
L.~Zhang, A.~Rao, and M.~Agrawala, ``Adding conditional control to text-to-image diffusion models,'' in \emph{Proceedings of the IEEE/CVF International Conference on Computer Vision (ICCV)}, 2023, pp. 3836--3847.

\bibitem{T2IAdapter}
C.~Mou \emph{et~al.}, ``{T2I-Adapter}: Learning adapters to dig out more controllable ability for text-to-image diffusion models,'' in \emph{Proceedings of the AAAI Conference on Artificial Intelligence (AAAI)}, vol.~38, no.~5, 2024, pp. 4296--4304.

\bibitem{LoRA}
E.~Hu \emph{et~al.}, ``{LoRA}: Low-rank adaptation of large language models,'' in \emph{International Conference on Learning Representations (ICLR)}, 2022.

\bibitem{CustomizeAVideo}
Y.~Ren \emph{et~al.}, ``{Customize-A-Video}: One-shot motion customization of text-to-video diffusion models,'' in \emph{European Conference on Computer Vision (ECCV)}, 2024, pp. 332--349.

\bibitem{FrameCrafter}
Q.~Wu, K.~Vuong, M.~Jeon, S.~Narasimhan, and D.~Ramanan, ``{FrameCrafter}: Novel view synthesis as video completion,'' in \emph{European Conference on Computer Vision (ECCV)}, 2026, pp. 73--91.

\bibitem{LiONLoRA}
Y.~Zhang, C.~Cao, C.~Yu, and J.~Zhu, ``{LiON-LoRA}: Rethinking {LoRA} fusion to unify controllable spatial and temporal generation for video diffusion,'' in \emph{Proceedings of the IEEE/CVF International Conference on Computer Vision (ICCV)}, 2025, pp. 14\,569--14\,579.

\bibitem{DDPM}
J.~Ho, A.~Jain, and P.~Abbeel, ``Denoising diffusion probabilistic models,'' \emph{Advances in Neural Information Processing Systems (NeurIPS)}, vol.~33, pp. 6840--6851, 2020.

\bibitem{Flow}
Y.~Lipman, R.~T.~Q. Chen, H.~Ben-Hamu, M.~Nickel, and M.~Le, ``Flow matching for generative modeling,'' in \emph{International Conference on Learning Representations (ICLR)}, 2023.

\bibitem{PFODE}
Y.~Song, J.~Sohl-Dickstein, D.~P. Kingma, A.~Kumar, S.~Ermon, and B.~Poole, ``Score-based generative modeling through stochastic differential equations,'' in \emph{International Conference on Learning Representations (ICLR)}, 2021.

\bibitem{DiffusionVisionSurvey}
F.-A. Croitoru, V.~Hondru, R.~T. Ionescu, and M.~Shah, ``Diffusion models in vision: A survey,'' \emph{IEEE Transactions on Pattern Analysis and Machine Intelligence}, vol.~45, no.~9, pp. 10\,850--10\,869, 2023, doi: 10.1109/TPAMI.2023.3261988.

\bibitem{Longcat}
Meituan LongCat Team \emph{et~al.}, ``{LongCat-Video} technical report,'' \emph{arXiv preprint arXiv:2510.22200}, 2025.

\bibitem{VGGT}
J.~Wang, M.~Chen, N.~Karaev, A.~Vedaldi, C.~Rupprecht, and D.~Novotny, ``{VGGT}: Visual geometry grounded transformer,'' in \emph{Proceedings of the IEEE/CVF Conference on Computer Vision and Pattern Recognition (CVPR)}, 2025, pp. 5294--5306.

\bibitem{UniDepth}
L.~Piccinelli \emph{et~al.}, ``{UniDepth}: Universal monocular metric depth estimation,'' in \emph{Proceedings of the IEEE/CVF Conference on Computer Vision and Pattern Recognition (CVPR)}, 2024, pp. 10\,106--10\,116.

\bibitem{OpticaFlow}
G.~Farneb{\"a}ck, ``Two-frame motion estimation based on polynomial expansion,'' in \emph{Scandinavian Conference on Image Analysis (SCIA)}, 2003, pp. 363--370.

\bibitem{RAFT}
Z.~Teed and J.~Deng, ``{RAFT}: Recurrent all-pairs field transforms for optical flow,'' in \emph{European Conference on Computer Vision (ECCV)}, 2020, pp. 402--419.

\bibitem{TeaCache}
F.~Liu \emph{et~al.}, ``Timestep embedding tells: It's time to cache for video diffusion model,'' in \emph{Proceedings of the IEEE/CVF Conference on Computer Vision and Pattern Recognition}, 2025, pp. 7353--7363.

\bibitem{DL3DV}
L.~Ling \emph{et~al.}, ``{DL3DV-10K}: A large-scale scene dataset for deep learning-based {3D} vision,'' in \emph{Proceedings of the IEEE/CVF Conference on Computer Vision and Pattern Recognition (CVPR)}, 2024, pp. 22\,160--22\,169.

\bibitem{Shapley1953}
L.~S. Shapley, ``A value for n-person games,'' in \emph{Contributions to the Theory of Games II}, ser. Annals of Mathematics Studies, H.~W. Kuhn and A.~W. Tucker, Eds.\hskip 1em plus 0.5em minus 0.4em\relax Princeton, NJ: Princeton University Press, 1953, vol.~28, pp. 307--317.

\bibitem{LLFF}
B.~Mildenhall \emph{et~al.}, ``Local light field fusion: Practical view synthesis with prescriptive sampling guidelines,'' \emph{ACM Transactions on Graphics (TOG)}, vol.~38, no.~4, Art. no.~29, pp. 1--14, 2019.

\bibitem{Tanks}
A.~Knapitsch, J.~Park, Q.-Y. Zhou, and V.~Koltun, ``Tanks and temples: Benchmarking large-scale scene reconstruction,'' \emph{ACM Transactions on Graphics (TOG)}, vol.~36, no.~4, Art. no.~78, pp. 1--13, 2017.

\bibitem{Mipnerf360}
J.~T. Barron, B.~Mildenhall, D.~Verbin, P.~P. Srinivasan, and P.~Hedman, ``{Mip-NeRF 360}: Unbounded anti-aliased neural radiance fields,'' in \emph{Proceedings of the IEEE/CVF Conference on Computer Vision and Pattern Recognition (CVPR)}, 2022, pp. 5470--5479.

\bibitem{DAVIS}
F.~Perazzi, J.~Pont-Tuset, B.~McWilliams, L.~Van~Gool, M.~Gross, and A.~Sorkine-Hornung, ``A benchmark dataset and evaluation methodology for video object segmentation,'' in \emph{Proceedings of the IEEE Conference on Computer Vision and Pattern Recognition (CVPR)}, 2016, pp. 724--732.

\bibitem{uniworld}
H.~Zhou, W.~Yu, C.~Feng, X.~Zhou, Y.~Tian, and L.~Yuan,
``UniWorld-View: Large-baseline view synthesis via video diffusion models,''
\emph{arXiv preprint arXiv:2608.04701}, 2026.

\bibitem{symphomotion}
G.~Zhang, Y.~Chen, X.~Xiang, J.~Huang, Z.~Wang, and L.~Jiang, ``{SymphoMotion}: Joint control of camera motion and object dynamics for coherent video generation,'' in \emph{Proceedings of the IEEE/CVF Conference on Computer Vision and Pattern Recognition (CVPR)}, 2026, pp. 11\,127--11\,137.

\bibitem{FID}
M.~Heusel, H.~Ramsauer, T.~Unterthiner, B.~Nessler, and S.~Hochreiter, ``{GANs} trained by a two time-scale update rule converge to a local {Nash} equilibrium,'' in \emph{Advances in Neural Information Processing Systems}, vol.~30, 2017, pp. 6626--6637.

\bibitem{CLIP}
A.~Radford \emph{et~al.}, ``Learning transferable visual models from natural language supervision,'' in \emph{Proceedings of the International Conference on Machine Learning (ICML)}, ser. Proceedings of Machine Learning Research, vol.~139, 2021, pp. 8748--8763.

\bibitem{FVD}
T.~Unterthiner, S.~van~Steenkiste, K.~Kurach, R.~Marinier, M.~Michalski, and S.~Gelly, ``Towards accurate generative models of video: A new metric \& challenges,'' \emph{arXiv preprint arXiv:1812.01717}, 2018.

\bibitem{lpips}
R.~Zhang, P.~Isola, A.~A.~Efros, E.~Shechtman, and O.~Wang, ``The unreasonable effectiveness of deep features as a perceptual metric,'' in \emph{Proceedings of the IEEE Conference on Computer Vision and Pattern Recognition (CVPR)}, 2018, pp. 586--595.

\bibitem{glomap}
L.~Pan, D.~Bar\'{a}th, M.~Pollefeys, and J.~L.~Sch\"{o}nberger, ``Global structure-from-motion revisited,'' in \emph{European Conference on Computer Vision (ECCV)}, 2024, pp. 58--77.

\bibitem{Pi3}
Y.~Wang \emph{et~al.}, ``$\pi^3$: Permutation-equivariant visual geometry learning,'' in \emph{International Conference on Learning Representations (ICLR)}, 2026.

\end{thebibliography}
\end{document}